\documentclass{bmvc2k}
\title{Are conditional GANs explicitly conditional?}
\addauthor{Houssem eddine Boulahbal$^1$}{houssem-eddine.boulahbal@renault.com}{2}
\addauthor{Adrian Voicila}{adrian.voicila@renault.com}{1}
\addauthor{Andrew I. Comport}{Andrew.Comport@cnrs.fr}{2}

\addinstitution{
 Renault Software Factory, \\
 2600 Route des Crêtes,\\
 06560 Valbonne, France
}
\addinstitution{
 CNRS-I3S, Université Côte d’Azur,\\
2000 Route des Lucioles,\\
06903 Sophia Antipolis, France
}

\runninghead{BOULAHBAL et al}{Are conditional GANs explicitly conditional?}
\def\eg{\emph{e.g}\bmvaOneDot}

\newcommand{\ac}{\textit{a~contrario} }
\newcommand{\act}{\textit{a-contrario} }
\newcommand{\mbf}[1]{{\mathbf{#1}}}
\newcommand{\be}{\begin{equation}}
\newcommand{\ee}{\end{equation}}

\newcommand{\beh}{\begin{equation*}}
\newcommand{\eeh}{\end{equation*}}

\newcommand{\bi}{\begin{itemize}}
\newcommand{\ei}{\end{itemize}}
\newcommand{\ba}{\begin{array}}
\newcommand{\ea}{\end{array}}
\newcommand{\bfg}{\begin{figure}}
\newcommand{\efg}{\end{figure}}
\newcommand{\bc}{\begin{center}}
\newcommand{\ec}{\end{center}}
\newcommand{\bbm}{\begin{bmatrix}}
\newcommand{\ebm}{\end{bmatrix}}
\usepackage{amsfonts}
\usepackage{xcolor}
\begin{document}
\maketitle
\begin{abstract}
 This paper proposes two important contributions for conditional Generative Adversarial Networks (cGANs) to improve the wide variety of applications that exploit this architecture. The first main contribution is an analysis of cGANs to show that they are not explicitly conditional. In particular, it will be shown that the discriminator and subsequently the cGAN does not automatically learn the conditionality between inputs. The second contribution is a new method, called \textbf{\ac~cGAN}, that explicitly models conditionality for both parts of the adversarial architecture via a novel \ac loss that involves training the discriminator to learn unconditional (adverse) examples. This leads to a novel type of data augmentation approach for GANs (\ac learning) which allows to restrict the search space of the generator to conditional outputs using adverse examples. Extensive experimentation is carried out to evaluate the conditionality of the discriminator by proposing a probability distribution analysis. Comparisons with the cGAN architecture for different applications show significant improvements in performance 
 on well known datasets including, semantic image synthesis, image segmentation, monocular depth prediction and "single label"-to-image using different metrics including Fréchet Inception Distance (FID), mean Intersection over Union (mIoU), Root Mean Square Error log (RMSE $\log$) and Number of statistically-Different Bins (NDB).
\end{abstract}

Since the seminal work in 2014, Generative Adversarial Networks (GANs)~\cite{goodfellow2014generative} have introduced an alternative framework for training generative models that has led to a multitude of high impact publications over a very large number of applications. Conditional GANs, introduced shortly after~\cite{mirza2014conditional}, have extended GANs to incorporate conditional information as input and have demonstrated resounding success for many computer vision tasks such as image synthesis~\cite{Isola2017ImagetoImageTW,Park2019SemanticIS,Wang2018HighResolutionIS,chen2017photographic,Sushko2020YouON,Tang2020LocalCA,Liu2019LearningTP}, video synthesis \cite{Wang2018VideotoVideoS,Chan2019EverybodyDN,Liu2019NeuralRA}, image correction\cite{Kupyn2018DeblurGANBM,Zhang2020ImageDU,Qu2019EnhancedPD}, text-to-image\cite{Reed2016GenerativeAT,zhang2018stackgan++,Xu2018AttnGANFT,Li2019ObjectDrivenTS}. In all these works, the underlying GAN model as proposed in~\cite{goodfellow2014generative} and~\cite{mirza2014conditional} have formed the basis for more advanced architectures and their properties have been analysed in detail and established in terms of convergence\cite{Kodali2018OnCA,Nie2018JRGANJR}, mode collapse\cite{Srivastava2017VEEGANRM}, Nash equilibrium\cite{Unterthiner2018CoulombGP,Farnia2020GANsMH}, vanishing gradients\cite{Arjovsky2017TowardsPM}, etc.

The « conditionality » of cGANs is at the crux of their theoretical contribution and its impact therefore merits in-depth analysis. 
From the existing literature it is not clear, however, if this now widely used architecture explicitly models or even learns conditionality. Empirically, the impressive results obtained with cGANs show that the generator automatically seeks to incorporate conditional variables into its generated output. Fundamentally the generator is, however, free to generate whichever output as long as it satisfies the discriminator. Therefore, the conditionality of cGAN also depends on the conditionality of the discriminator. This begs the question as to whether or not the baseline architecture of cGANs explicitly models conditionality and if not, how can the core adversarial architecture be redefined to explicitly model conditionality? This is therefore the object of this paper.

Problems with cGANs conditionality have been observed independently for different tasks in the literature. Label-to-image tasks observe that using only adversarial supervision yields bad quality results~\cite{Sushko2020YouON,Park2019SemanticIS, Wang2018HighResolutionIS,Liu2019LearningTP}. "Single Label"-to-image tasks~\cite{Brock2019LargeSG} observes class leakage. It is also well known that cGANs are prone to mode collapse~\cite{richardson2018gans}. In this paper it is suggested that all these problems are related to the lack of a conditional discriminator.

\begin{figure*}
    \centering
    \includegraphics[width=0.9\textwidth]{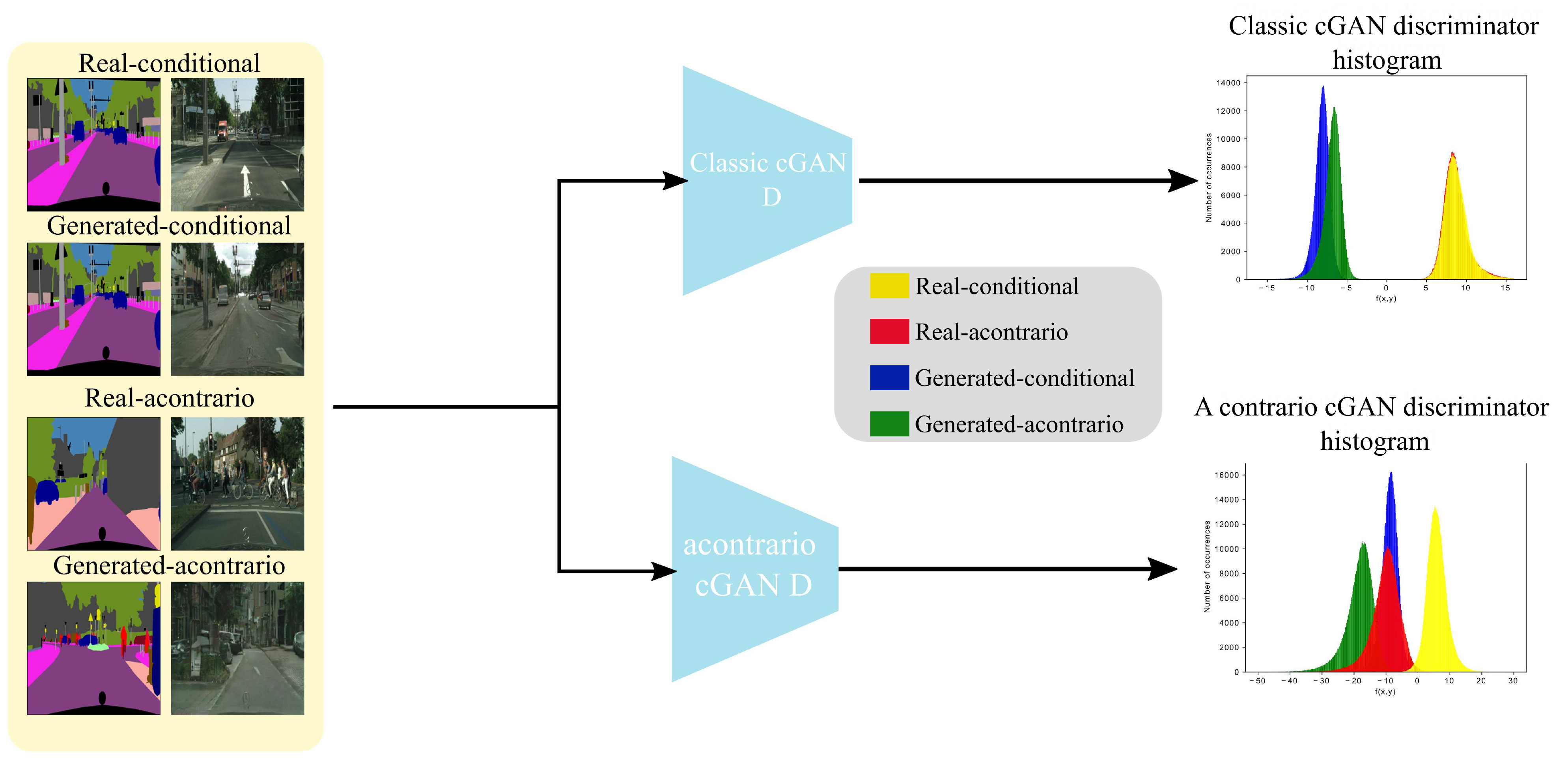}
    \caption{\small	 The classic cGAN and the proposed \ac cGAN discriminators are tested with 500 validation images of the Cityscapes dataset on both conditional and unconditional label-to-image input set.  Unconditional inputs (Real \ac and Generated \ac) set are obtained by randomly shuffling the original conditional sets of data. The classic cGAN discriminator fails to classify unconditional input set as false as seen by the histogram distributions on the right (real \ac in red is classified as true). The proposed method trains the discriminator with a general \ac loss to classify unconditional input set as fake (note that no extra training samples are required). The he proposed \ac cGAN correctly classifies all four modalities (blue, green, red, yellow) correctly.}
    \label{fig:pipeline}
\end{figure*}
Consider a simple test of conditionality on a learnt discriminator for the task of label-to-image translation, shown in Figure~\ref{fig:pipeline}. The conditional label input is purposely swapped with a non-corresponding input drawn randomly from the input set (\eg labels). From this test it is revealed that the discriminator does not succeed to detect the entire set of \ac examples (defined in Section~\ref{sec:consistency}) as false input pairs. More extreme cases are shown in~\ref{fig:conditionality}. This suggests that the generator is not constrained by the discriminator to produce conditional output but rather to produce any output from the target domain (images in this case). Furthermore, in practice, the large majority of methods that exploit cGANs for label-to-image translation, if not all, add additional loss terms to the generator to improve conditionality. These loss terms are, however, not adversarial in their construction. For example, high resolution image synthesis approaches such as~\cite{Wang2018HighResolutionIS} suffer from poor image quality when trained with only adversarial supervision~\cite{Sushko2020YouON}. Considering the well known pix-to-pix architecture~\cite{Isola2017ImagetoImageTW}, a L1 loss was introduced to improve performance. This additional term seeks to enforce conditionality on the generator but does not act explicitly on the discriminator. Subsequently, one could question if the conditionality obtained by such methods is obtained via this loss term which is not part of the adversarial network architecture. Moreover, adding an extra loss term to the generator has now become the defacto method for improving cGANs results. For example perceptual loss~\cite{Johnson2016PerceptualLF} and feature-matching~\cite{Salimans2016ImprovedTF} have been proposed and reused by many others\cite{Wang2018HighResolutionIS,chen2017photographic,Park2019SemanticIS,Ntavelis2020SESAMESE}. As demonstrated in the experiments different tasks such as image-to-depth or image-to-label also exhibit these drawbacks.

In this paper it will be argued that simply providing condition variables as input is insufficient for modelling conditionality and that it is necessary to explicitly enforce dependence between variables in the discriminator. It will be demonstrated that the vanilla cGAN approach is not explicitly conditional via probabilistic testing of the discriminator's capacity to model conditionality.
With this insight, a new method for explicitly modelling conditionality in the discriminator and subsequently the generator will be proposed. This new method not only offers a solution for conditionality but also provides the basis for a general data augmentation method by learning from the contrary (\ac data augmentation).

\section{Related work}

\paragraph*{Conditional GANs}
Generative adversarial networks~\cite{goodfellow2014generative} can be considered as a zero-sum game between two player neural networks $G$ and $D$ competing to reach a Nash equilibrium. This game is commonly formulated through a min-max optimization problem as follows:
    \begin{align}
        \min_{G\in\mathbb{G}}\max_{D\in\mathbb{D}} V(G,D) \label{eq:cost}
    \end{align}
where $\mathbb{G}$ and $\mathbb{D}$ are respectively the generator and discriminator function sets. A GAN is considered conditional~\cite{mirza2014conditional} when the generator's output is conditioned by an extra input variable. The condition variable can be any kind of information such as a segmentation mask, depth map, image or data from other modalities. 

There are various methods that have been proposed for incorporating conditional information into the generator~\cite{Vries2017ModulatingEV,Dumoulin2017ALR,Zhou2020SearchingTC,Zhou2020SearchingTC,Huang2017ArbitraryST,Park2019SemanticIS}. Recently~\cite{Tang2020LocalCA} introduced a classification-based feature learning module to learn more discriminating and class-specific features. Additional generator losses have also been proposed including feature matching~\cite{Salimans2016ImprovedTF}, perceptual loss~\cite{Johnson2016PerceptualLF} and cycle-consistency loss~\cite{zhu2017unpaired}. While all these methods improve the conditionality of the generator, they assume that the discriminator models conditionality. 

Alternatively several methods have been proposed to incorporate conditional information into the discriminator. \cite{mirza2014conditional} proposed an early fusion approach by concatenating the condition vector to the input of the discriminator.  \cite{Miyato2018cGANsWP,Kang2020ContraGANCL,Ntavelis2020SESAMESE,Liu2019LearningTP,kavalerov2021multi} proposed a late fusion by encoding the conditional information and introducing it into the final layers of the discriminator. \cite{Sushko2020YouON} replaces the discriminator with a pixel-wise semantic segmentation network. Most of these methods are task specific and not general~\cite{kavalerov2021multi,Sushko2020YouON,Kang2020ContraGANCL}. The method proposed in this paper is a task agnostic solution for incorporating any conditional variables and it will be shown to perform on several different datasets.

One concern raised in the literature is that improving either player is at the detriment to learning and can lead to mode collapse or vanishing gradient~\cite{Arjovsky2017TowardsPM, dai2017good}. In the present paper, whilst the aim is to define a conditional discriminator, it is argued that not only this does not affect the Nash equilibrium but also helps to avoid mode collapse and vanishing gradient. A simple example is that the proposed discriminator imposes conditionality and therefore the different condition inputs cannot be collapsed into a single mode.

\paragraph*{Data augmentation for GANs}
Data augmentation is a highly useful strategy to improve generalization for machine learning that involves increasing the diversity of training data without collecting new samples. Augmentation strategies have involved performing a variety operations on input training data including random cropping, horizontal flipping, various hand-crafted operations~\cite{hendrycks2019augmix,ghiasi2020simple} or even learning augmentation strategies from data~\cite{zhang2019adversarial,cubuk2019autoaugment}. Performing data augmentation on GANs is less straightforward and has been limited to a handful of operations such as flipping, cropping and jittering~\cite{Isola2017ImagetoImageTW}.  \cite{Zhao2020ImprovedCR,Zhang2020ConsistencyRF} propose a method to enforce the discriminator to remain unchanged by arbitrary augmentation. \cite{zhao2020image,zhao2020differentiable,Karras2020TrainingGA,9319516} study the artifacts produced by augmentation and propose to perform augmentation on both real and generated images for both generator and discriminator training. Further, \cite{Karras2020TrainingGA} proposes a method called stochastic discriminator augmentation that adjusts the augmentation strength adaptively. Alternatively to all these approaches, the \ac learning proposed in this paper performs data augmentation by providing false counter examples rather than augmenting the real data-set with transformations. The proposed data augmentation approach is therefore specific to adversarial architectures, however, it will be shown that by discriminating against undesirable distributions, the generator is better constrained towards the target distribution. In the specific case of conditionality, by providing \ac examples to the discriminator as fakes, the discriminator constrains the generator to only learn conditional generations and the generator is therefore better constrained. It should be noted that in parallel to this work, the idea of augmenting the discriminator with fake examples was recently proposed in~\cite{sinha2020negative}, however, this method did not consider discriminator conditionality and only investigated data augmentation through creating fake data.  

\paragraph*{Evaluation criteria for GANs}
\label{sec:related_work_eval}
Defining an evaluation criteria to assess and compare GAN architectures has been a major problem in advancing the field. The difficulty and advantage of GAN architectures lies in that the learnt discriminator replaces the classic objective function. 
Classic approaches for evaluating GANs often use the FID~\cite{Heusel2017GANsTB} or Inception scores~\cite{Salimans2016ImprovedTF} to evaluate high-level generation quality and diversity. For example, the inception score only allows to evaluate a perceptual image quality and does not account for consistency between generated output and label. By construction it is therefore domain specific and does not evaluate conditionality in the output. More recent approaches investigate more general metrics~\cite{arora2017generalization,huang2017parametric,im2018quantitatively,gulrajani2020towards,zhang2018unreasonable}. Nevertheless, none of these approaches specifically allow to evaluate the conditionality of conditional GANs. Such an evaluation metric should take both the condition variables and generated variables into account.
Therefore, for the sake of evaluation, experiments were done on tasks that could provide a strong conditional metric to evaluate conditionality. The mIoU was used specifically for evaluating semantically conditional examples and $\log$ RMSE was considered for geometric depth conditionality.

\section{Method}
\label{sec:method}

\subsection{Classic cGAN}
\label{sec:ccgan}

Classical cGAN training is based on conditionally paired sets of data $\mathcal{C}(\mbf{x},\mbf{y})$ where $\mbf{x}\sim p(\mbf{x})$ is the condition variable and $\mbf{y}\sim p(\mbf{y}|\mbf{x})$ is the real training variable. The generator of a cGAN outputs a transformed set of data $\mathcal{C}_G(\mbf{x},\mbf{y}_G)$ composed of the generator output variable $\mbf{y}_G\sim p_G(\mbf{y})$ and the condition variable. These sets of data will be called "real-conditional" and "generated-conditional" respectively. 
The discriminator is defined as:
\begin{align}
    D(\mbf{x},\mbf{y}):=\mathcal{A}(f(\mbf{x},\mbf{y}))
    \label{eq:discriminator}
\end{align}
Where $f(.)$ is a neural network function of $\mbf{x}$ and $\mbf{y}$, and $\mathcal{A}$ is the activation function whose choice depends on the objective function. The cGAN objective function is defined as: 
\begin{align}
         \mathcal{L}_{adv}= \min_{G} \max_{D} \left(\: \mathbb{E}_{\mbf{x}\sim p(\mbf{x}),\mbf{y}\sim p(\mbf{y}|\mbf{x})}\big[log(D(\mbf{x},\mbf{y})]\big] +   \mathbb{E}_{\mbf{x}\sim p(\mbf{x})}\big[log[1 - D(\mbf{x},G(\mbf{x}))]\big] \right)
\label{eq:classic_loss}
\end{align}
The min-max activation function is defined as a Sigmoid $\mathcal{A}(\mbf{x})=\left({\frac{1}{1+e^{-\mbf{x}}}}\right)$. 

\subsection{Evaluating conditionality} 
\label{sec:conditioneval}
The objective of this section is to propose methods to test the conditionality of cGAN networks. As mentioned in Section~\ref{sec:related_work_eval}, state-of-the-art approaches have focused on evaluating cGAN architectures with metrics applied to the generator output. Since the generator and discriminator are coupled, these metrics essentially evaluate the full GAN architecture.

A proposal is made to test the conditionality by visualizing the probability distribution at the output of the discriminator. Due to the fact that adversarial training involves a zero-sum game between a generator and a discriminator, both the generator and discriminator should seek to reach an equilibrium (Eq~\ref{eq:cost}) at the end of training. One issue for GANs is that when the discriminator dominates there is a vanishing gradient problem~\cite{Arjovsky2017TowardsPM}. It is therefore more difficult (but not impossible) to isolate the discriminator during training to evaluate its capacity to detect unconditional examples as false. For this reason, an optimal discriminator can be used to give insight for evaluation purposes as in~\cite{Arjovsky2017TowardsPM,Farnia2020GANsMH}. An optimal discriminator is essentially a binary classifier which classifies between true and fake (see Eq~\eqref{eq:classic_loss}).

In order to test the optimal discriminator, consider that the generator has been fixed after a certain number of iterations (the generator is initially random) and the discriminator has been allowed to converge to an optimal solution based on the following objective function:
 \begin{align}
    \max_{D\in \mathbb{D}} V(G_{fixed},D) 
    \label{eq:optimaldiscrim}
\end{align}

The evaluation subsequently involves analysing the distributions produced by the optimal discriminator (Eq~\eqref{eq:optimaldiscrim}) given test distributions containing unconditional or \ac sets of data-pairings. The capacity of the discriminator to correctly classify unconditional data as false is then analysed statistically. Section~\ref{sec:consistency} provides a formal definition of these unconditional data pairings. Probability distributions are visualised and evaluated by histogram analysis on the discriminator features in the last convolution layer. 
Refer to Section~\ref{sec:histeval} for a more theoretical analysis and detailed results.

\subsection{A contrario conditionality loss}
\label{sec:consistency}

The proposed \ac cGAN approach is based on training with unconditionally paired sets of data, obtained by randomly shuffling or re-paring the original conditional sets of data. The \ac set is defined as $\mathcal{C}_U(\mbf{\tilde x},\mbf{y})$, where $\mbf{\tilde x}\sim p(\mbf{x})$ is the \ac conditional variable ($\mbf{\tilde x} \neq \mbf{x}$) and $\mbf{y}$ is the real training variable as in Section~\ref{sec:ccgan}. In this case $\mbf{\tilde x}$ and $\mbf{y}$ are independent. The generator of the \ac cGAN outputs a transformed set of data $\mathcal{C}_{UG}(\mbf{\tilde x},\mbf{y})$ composed of the generator output variable $\mbf{y}_G\sim p_G(\mbf{y})$ and the random variable $\mbf{\tilde x}$. For the purpose of this paper these two sets of data will be called "real-\ac" and "generated-\act" respectively. The motivation to create these new sets is to train the discriminator to correctly classify unconditional data as false. Figure~\ref{fig:pipeline} shows the four possible pairings. In practice, random sampling of \ac pairs is carried out without replacement and attention is paid to not include any conditional variables into a same batch while processing.

In order to enforce conditionality between $\mbf{y}$ and $\mbf{x}$ an \ac term is proposed as:
\begin{align}
        \mathcal{L}_{ac}= \max_{D} \: & \left(\mathbb{E}_{\mbf{\tilde x} \sim p(\mbf{\tilde x}) ,\mbf{y}\sim p(\mbf{y})}\big[log(1 - D(\mbf{\tilde x},\mbf{y}))\big]  + 
        \mathbb{E}_{\mbf{\tilde x} \sim p(\mbf{\tilde x}) ,\mbf{x} \sim p(\mbf{x})}\big[log(1 - D(\mbf{\tilde x},G(\mbf{x})))\big] \right)
\label{eq:ac_loss}
\end{align}
The first term is enforces the real-\act pairs to be classified as fakes. The second terms enforces the generated-\act as fake. The final loss is:
\begin{align}
    \mathcal{L}_{adv}^{'}  = \mathcal{L}_{adv}+ \mathcal{L}_{ac}
\label{eq:overall_loss}
\end{align}

\linespread{0.86}
\section{Experimental section}
\label{sec:experiments}
Several experiments will be presented that evaluate the conditionality of cGANs including: Real image generation from semantic masks on Cityscapes dataset~\cite{Cordts2016Cityscapes};"Single label"-to-image on CIFAR-10~\cite{cifar}; Monocular depth estimation on~\cite{Silberman:ECCV12}; Semantic segmentation using pix2pix on Cityscapes dataset. 
For label-to-image generation, pix2pix\cite{Isola2017ImagetoImageTW},  pix2pixHD\cite{Wang2018HighResolutionIS}, SPADE\cite{Park2019SemanticIS} and CC-FPSE\cite{Liu2019LearningTP} were used to test the conditionality and to highlight the contribution of the \ac cGAN wrt state-of-the-art approaches. "Single label"-to-image is also considered as most new techniques that improves cGANs are designed tested using this task. To demonstrate the generality of the proposed approach depth estimation and image-to-label segmentation tasks are also performed. These structured prediction problems offer strong metrics for evaluating cGANs and various public datasets are available for training. While the scope of conditional evaluation has been limited to tasks that could provide a metric to evaluate both the conditionality and the quality of the generation, the proposed approach is general and not specific to these particular tasks. During training, the network's architecture, the additional losses, the hyper-parameters
and data augmentation schemes are kept as in the original papers ~\cite{Isola2017ImagetoImageTW, Park2019SemanticIS, Wang2018HighResolutionIS, Liu2019LearningTP}. The new additional \ac term is the only difference between the compared methods. 

\subsection{Evaluating conditionality}
\label{sec:eval}
\begin{figure*}
\centering    \subfigure{\includegraphics[width=0.245\textwidth]{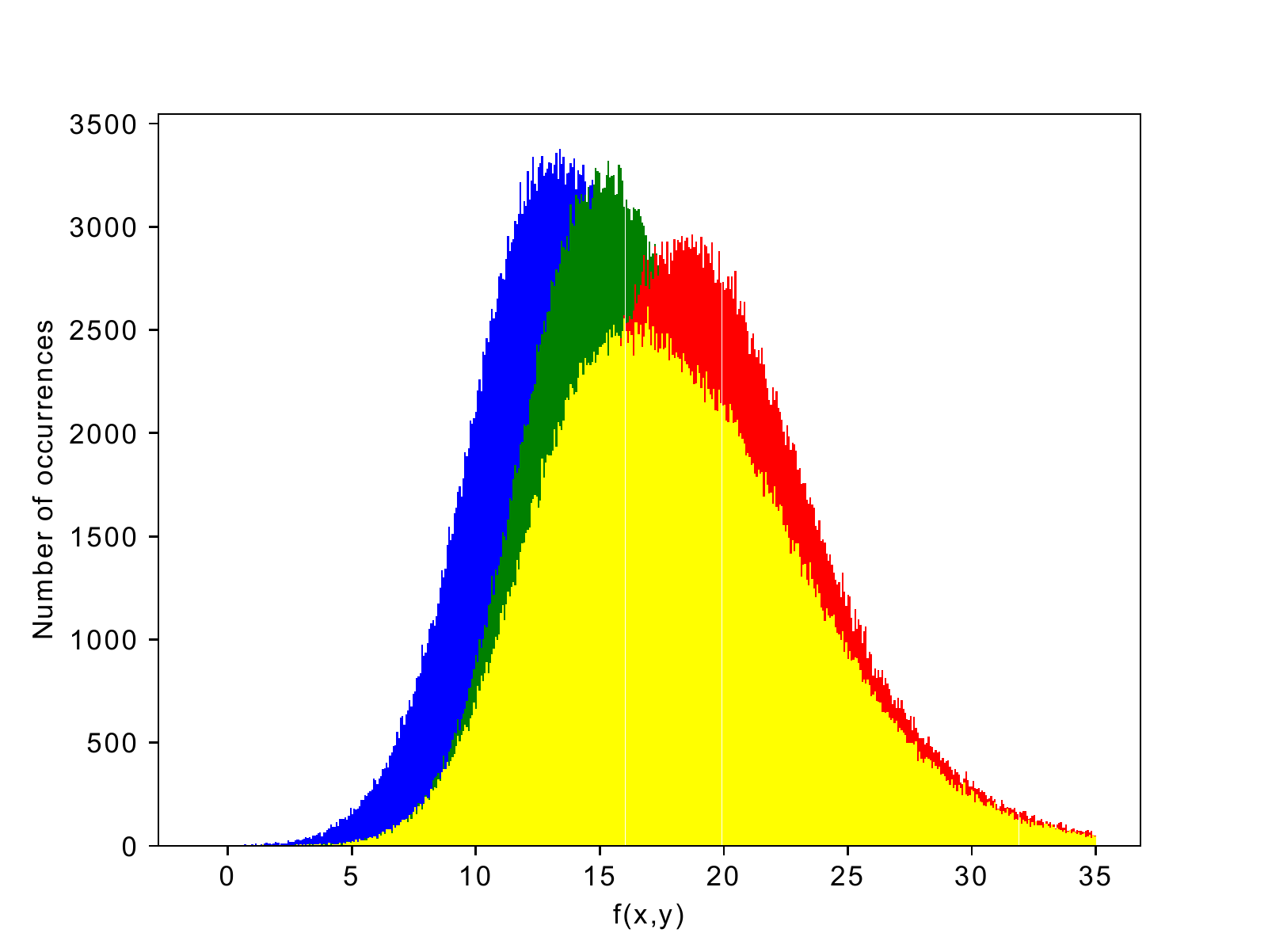}} 
    \subfigure{\includegraphics[width=0.245\textwidth]{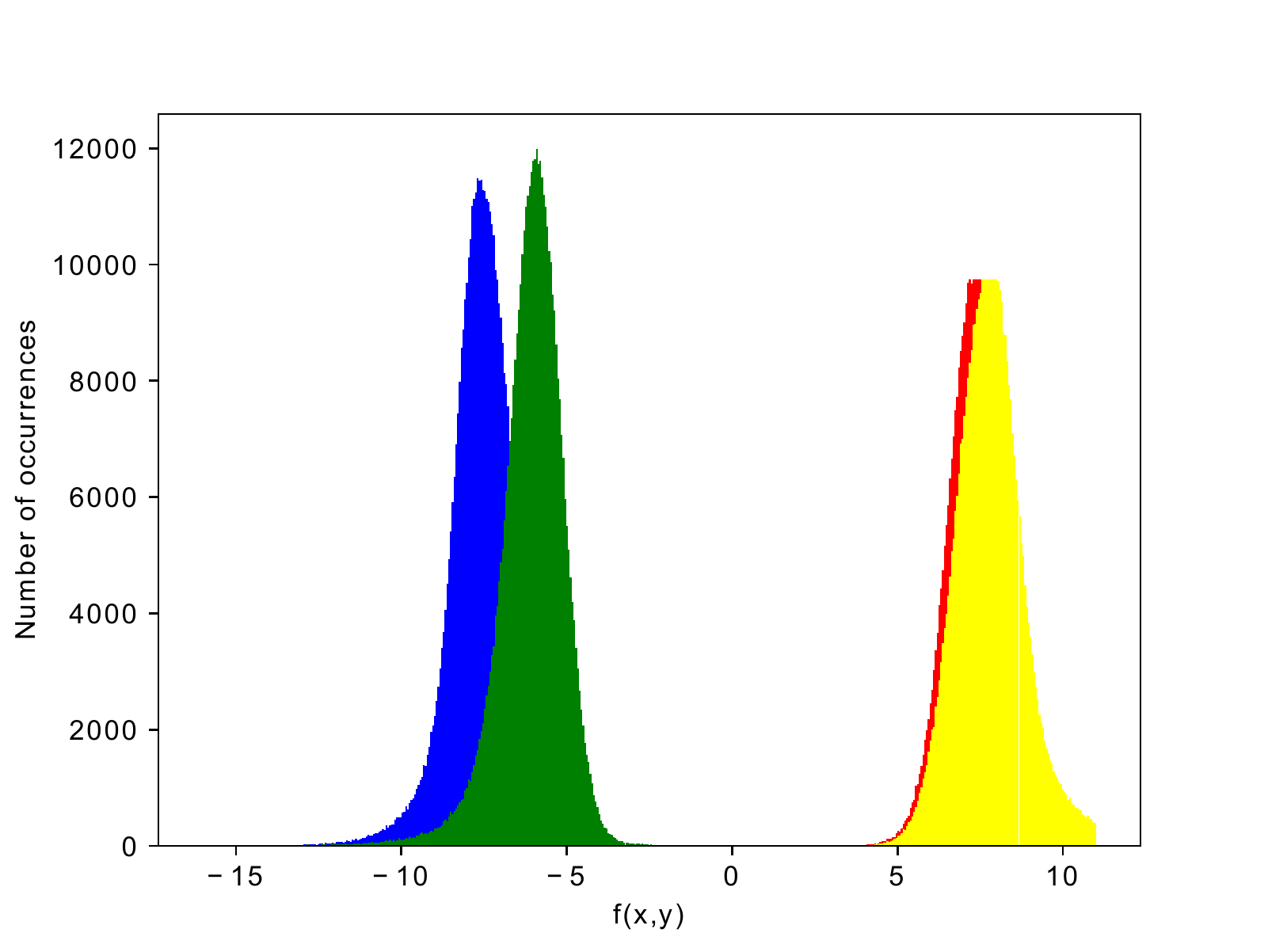}} 
    \subfigure{\includegraphics[width=0.245\textwidth]{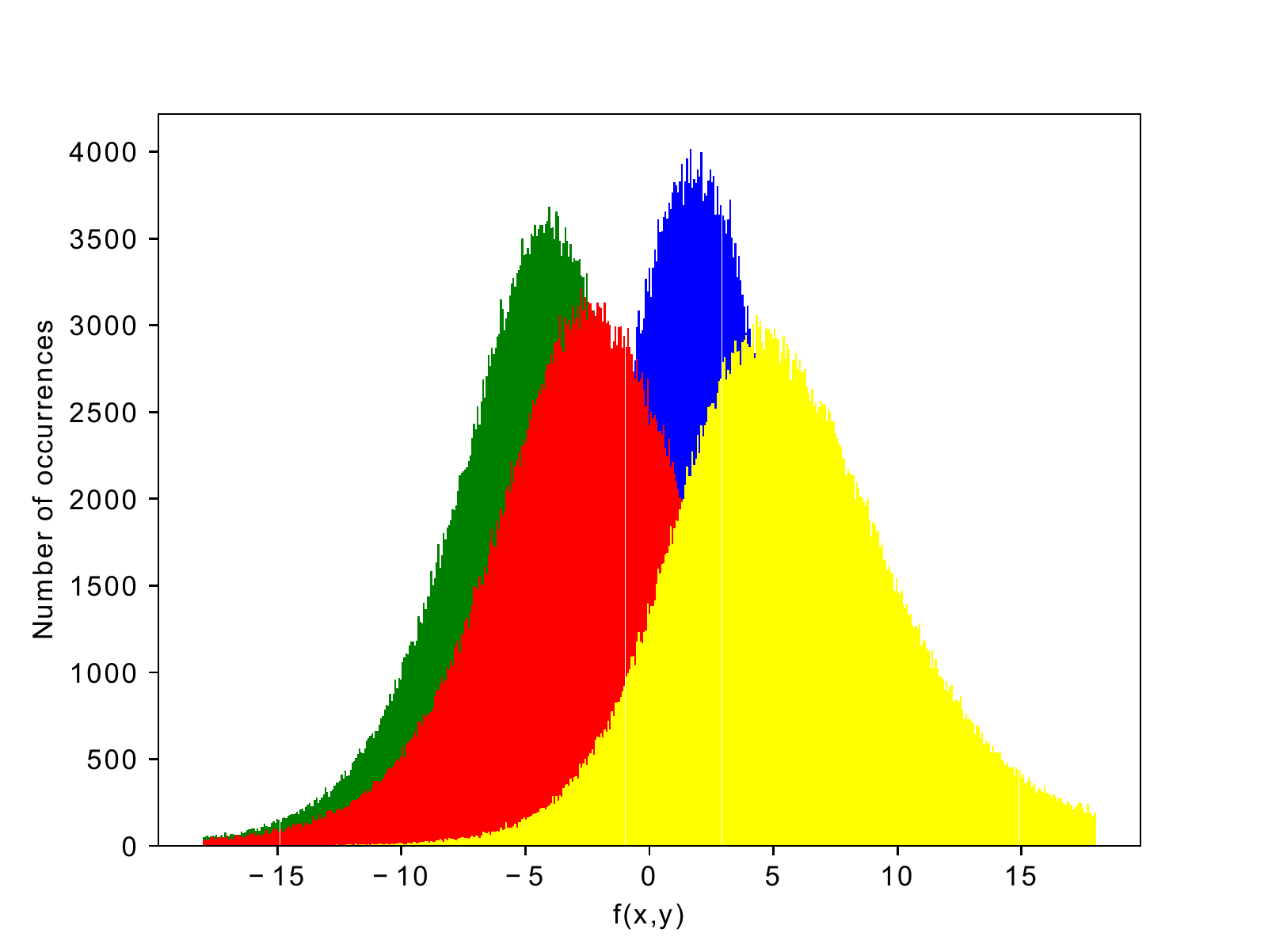}}
     \subfigure{\includegraphics[width=0.245\textwidth]{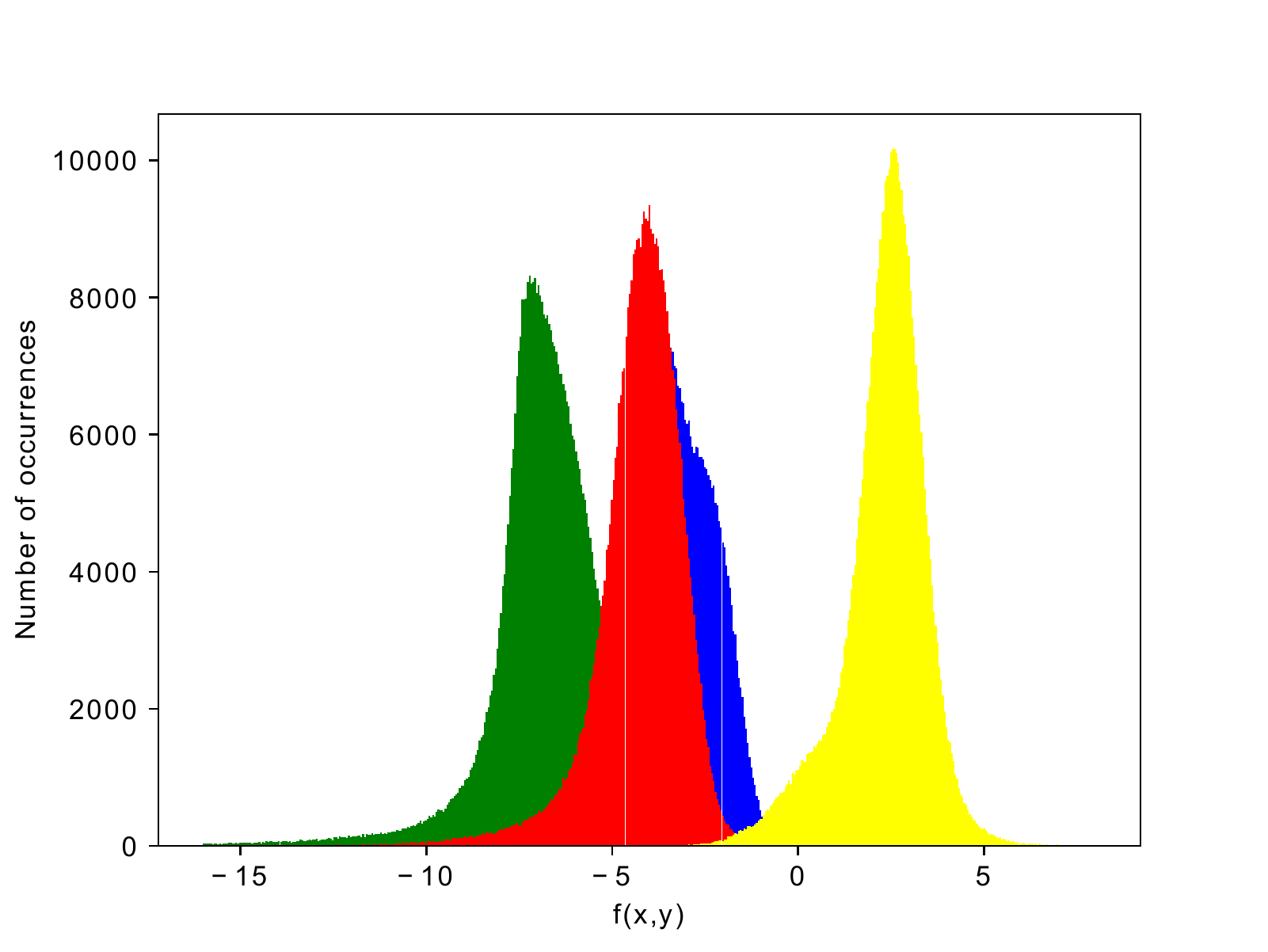}} 
    \caption{\small	Label-to-image histogram results when validating 500 Cityscape images on a discriminator trained until epoch 200. Blue is Generated-conditional, Green is generated \ac, Red is real-\ac, Yellow is Real-conditional. (a) The trained baseline discriminator, (b) Optimal baseline discriminator, (c) \ac cGAN discriminator, (d) Optimal \ac cGAN discriminator. (a) and (c) are still learning indicating no vanishing gradient or mode collapse~\cite{Arjovsky2017TowardsPM}. (b) doesn't detect conditionality since \ac real is classified as true (red) (d) succeeds to classify all modes correctly.}
    \label{fig:dist}  
\end{figure*}
Preliminary conditionality evaluation follows the method presented in Section~\ref{sec:conditioneval} using \ac sets to evaluate an optimal discriminator. 
In a first part experiments were carried out on the vanilla pix2pix cGAN with a discriminator PatchGAN architecture with $70\times70$ receptive field. The model was trained on the Cityscapes dataset~\cite{Cordts2016Cityscapes} for label-to-image translation with 2975 training images resized to $256\times256$. The pix2pix GAN model is trained with the same hyper-parameters as specified in the original paper~\cite{Isola2017ImagetoImageTW}. The evaluation histogram is calculated on the values of the last convolution layer of the discriminator ($f(x,y)$ of Eq~\eqref{eq:discriminator}) based on the $500$ validation images. Each sample from the last convolutional layer is composed of a $30\times30$ overlapping patches with one channel.The proposed approach is trained in exactly the same manner with the only difference being the new objective function. 

Various tests were carried out to investigate the output distributions of each set of data for both the baseline architecture and the proposed method. The underlying accuracy of the implementation was first validated to ensure the accuracy reported in the original paper.  A histogram analysis was then performed for different levels of training including: training for 20, 100, 200 epochs and evaluating after each. In another experiment the discriminator was allowed to continue to converge for one epoch after fixing 20, 100 and 200 epochs of cGAN training. In particular, training is performed with the objective given in Eq~\eqref{eq:optimaldiscrim} and as proposed in~\cite{Arjovsky2017TowardsPM, Farnia2020GANsMH}. These results are plotted for each data pairing: real-conditional,  generated-conditional, real-\act and generated-\act in Figure~\ref{fig:dist}. 
    Figure~\ref{fig:dist} (a) and (c) show that during training the generator is still learning with no vanishing gradient or mode collapse. In Figure~\ref{fig:dist} (b) the discriminator has been allowed to reach an optimal value by fixing the generator. The real \ac pairing is wrongly classified $99.9\%$ of the time indicating that the discriminator has not learnt conditionality. (d) Shows clearly four distinct distributions and shows the ability of the proposed approach to learn conditionality and correctly classify real \ac pairing $91.9\%$ of the time. Refer to Section~\ref{sec:histeval} for more insight into the histogram evaluation of conditionality. Similar conditionality tests were performed for various alternative architectures including using a separate/shared network for $\mbf{x}$ and $\mbf{y}$ and early/late/at-each-layer fusion. In all cases conditionality was not learnt.

The same analysis was performed for the monocular depth prediction dataset and similar observations were made (see Figure~\ref{fig:histograms-appendix}). 
These results strongly suggest that classic cGAN is unable to learn conditionality and that the spectacular results obtained by cGAN architectures are largely due to higher a level style constraints that are not specific to the input condition variable since swapping condition variables produces no effect. The proposed histogram test allows to demonstrate the ability of the discriminator to classify the various underlying classes of data and shows their statistical distribution. The approach proposed in~\cite{Arjovsky2017TowardsPM, kavalerov2021multi}
 looks only at accuracy, while the proposed histogram test also allows to see how well the discriminator distinguishes between the different data modalities (real/fake/\ac).

\subsection{Label-To-Image translation}
\label{sec:lb-to-im}
\begin{figure*}
\centering    
\subfigure{\includegraphics[width=0.30\textwidth]{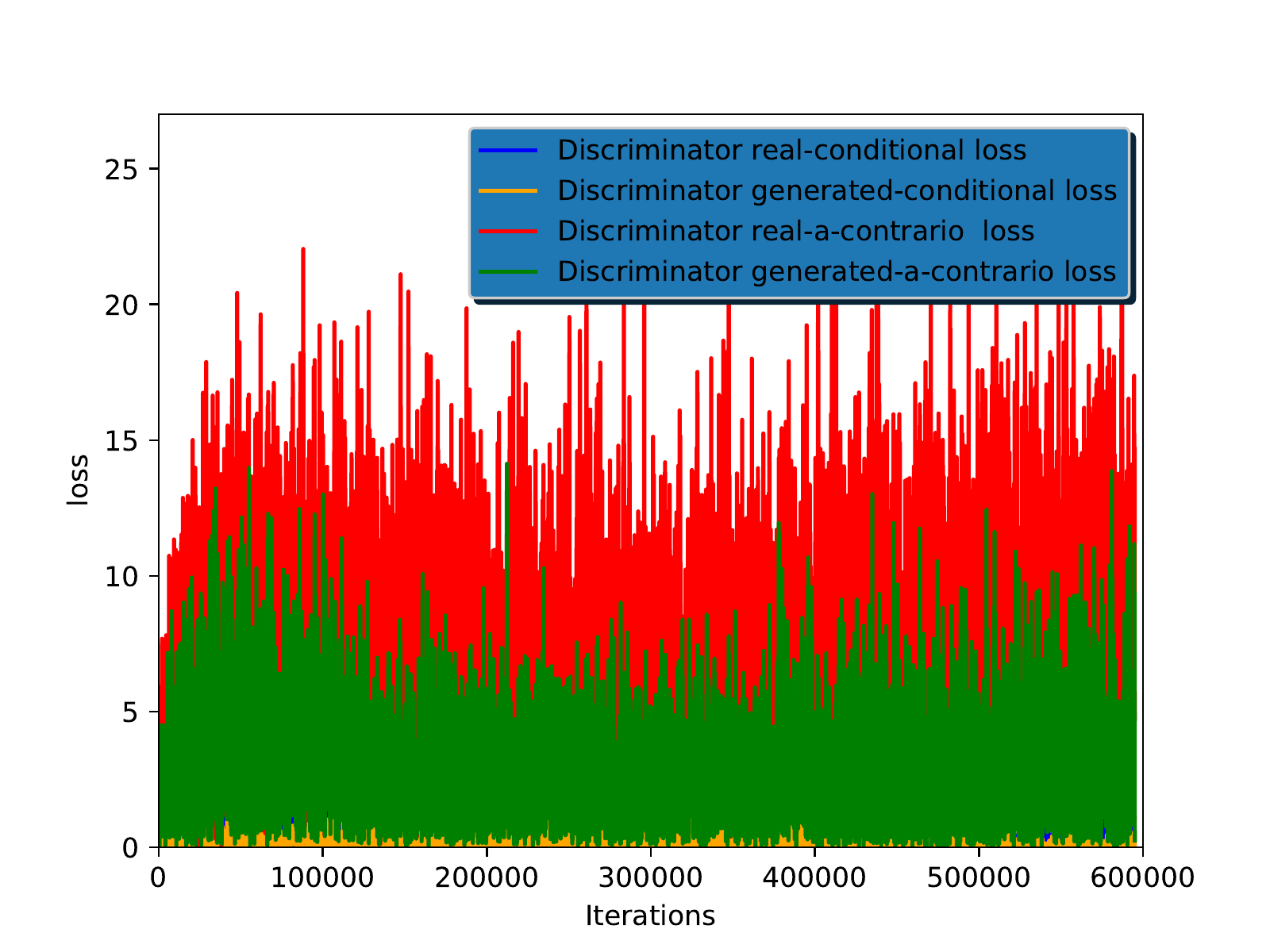}}
\subfigure{\includegraphics[width=0.30\textwidth]{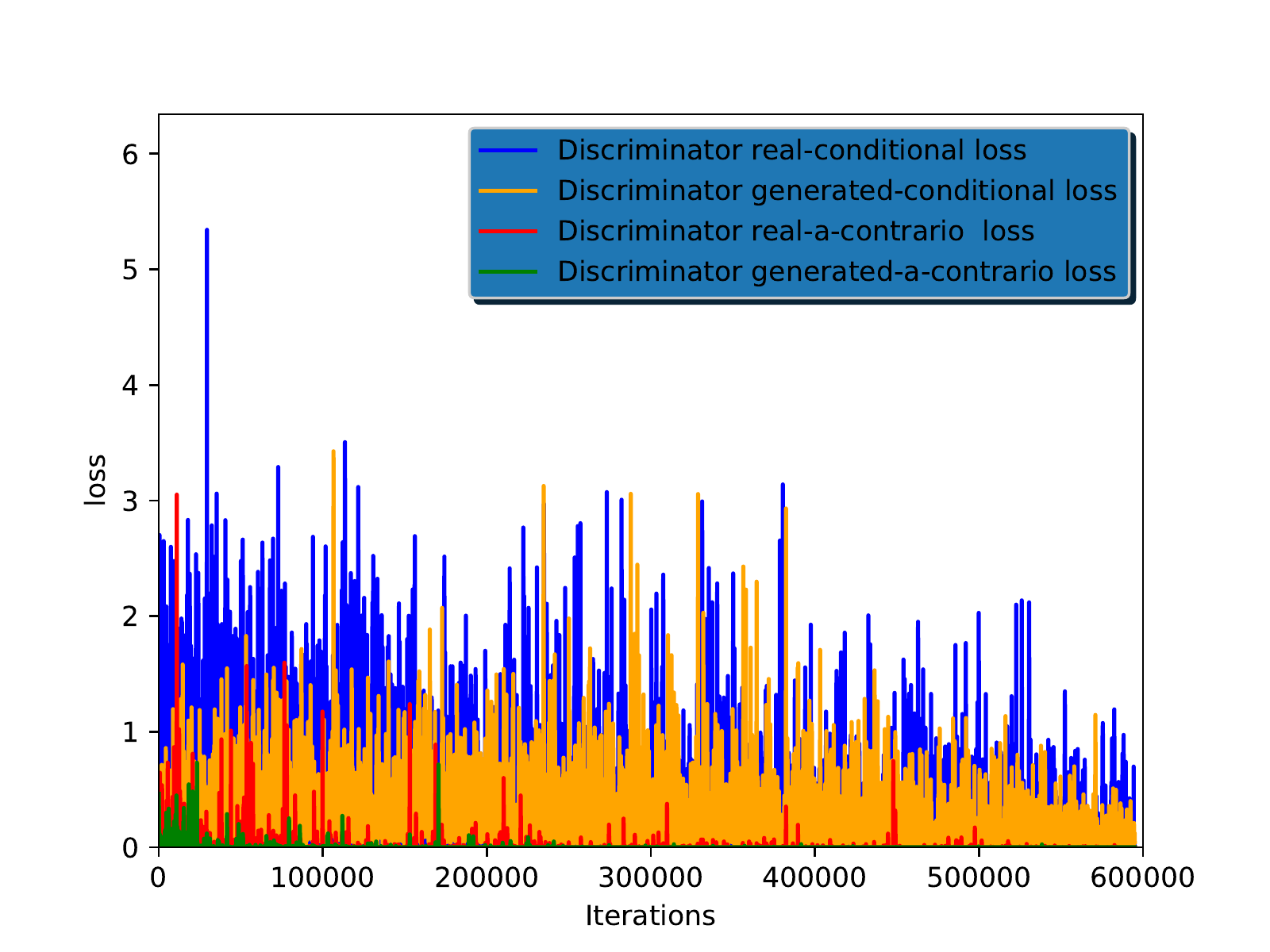}}
\subfigure{\includegraphics[width=0.30\textwidth]{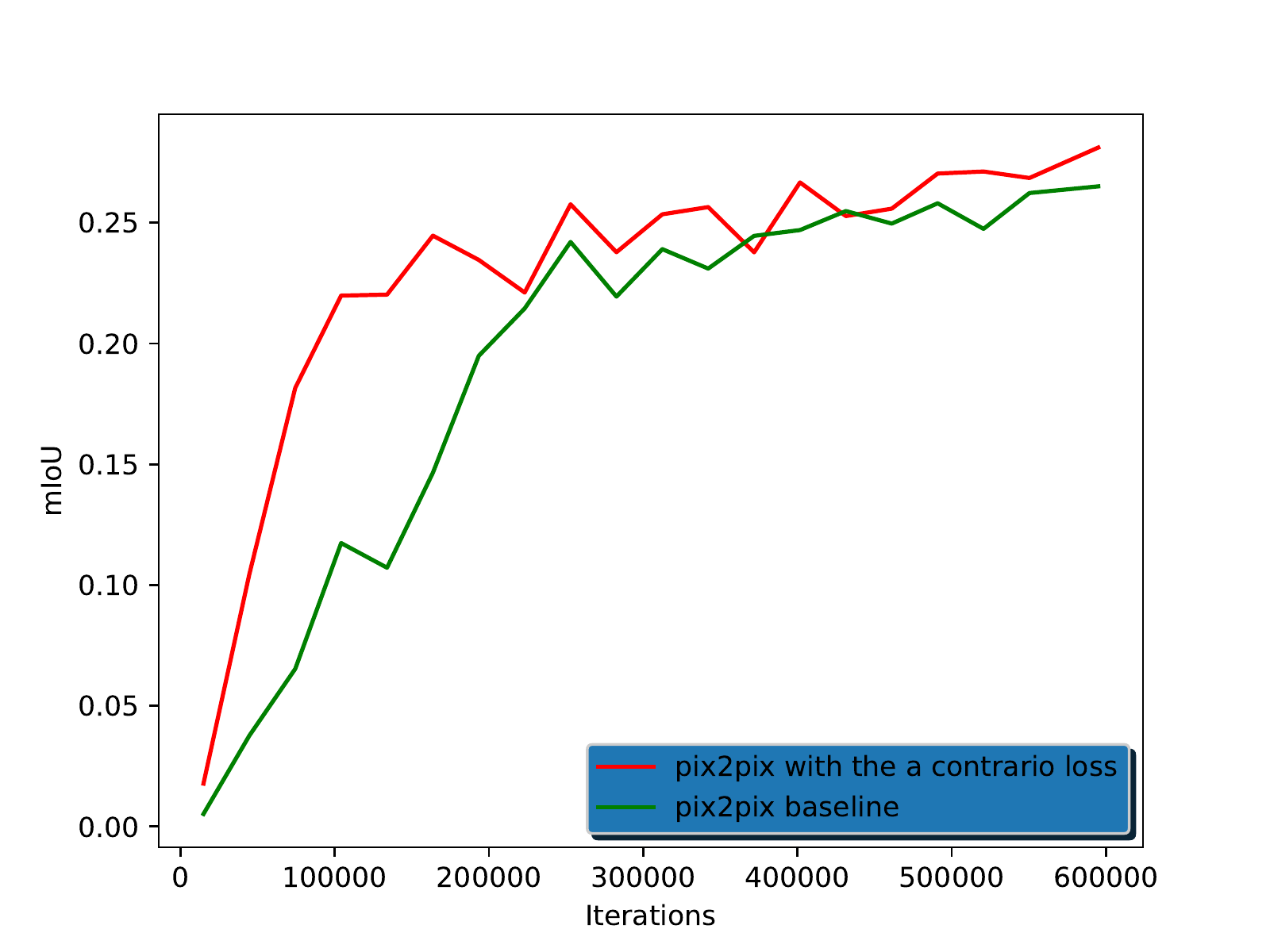}}
    \caption{\small	Comparison of the proposed approach on the Cityscape label-to-image training set. (a) The loss function for each set of data-pairing for the baseline cGAN method (\ac are for evaluation only). (b) The loss function for each set of data-pairing for the \ac cGAN method. (c) The evolution of the mIOU for both methods, performed on the validation dataset. It can be seen in (b) compared to (a) that the \ac loss converges to 0 rapidly for the proposed approach. In (c) the proposed approach is much more efficient and converges much faster and with higher accuracy.}
    \label{fig:mIoU}
\end{figure*}
 Generating realistic images from semantic labels is good task to evaluate the effect of the \ac at a high level since many images can be potentially generated for each semantic class label. Figure~\ref{fig:mIoU}(c) shows a comparison of the mIoU for the baseline pix2pix model and proposed pix2pix model with the additional \ac loss. It can be observed that the \ac cGAN converges faster than the baseline. The mIoU of the model with \ac at iteration $163$k is $24.46$ whereas the baseline is $14.65$. The mIoU oscillates around that value for the \ac model indicating that the model has converged. After $595$k iterations, the mIoU for both models are very close $28.28$ and $26.41$. It is worth noting that evaluating using real images yields $29.6$. The convergence is reported for the generator where the computational cost is exactly the same. the \ac loss is specific only to the discriminator and adds a small computational cost. By restricting the search space of the generator to only conditional pairs, the generator's convergence is faster.

Table~\ref{tab:high_resolution} shows a comparison of different architectures with and without \ac augmentation. For a fair comparison all the networks are trained from scratch and the same hyper-parameter are used. The \ac loss is the only difference between the two networks. The batch size for SPADE is 32 and 16 for CC-FPSE. Through explicitly enforcing the conditionality with \ac examples, the discriminator learns to penalize unconditional generation achieving better results.  Figure~\ref{fig:state_of_art_comparison} shows a qualitative comparison. 

Moreover, Figure~\ref{fig:mIoU}(a) and Figure~\ref{fig:mIoU}(b) show the comparison of the losses of the discriminator for both models on this dataset. The baseline is trained with only conditional pairs, however, the \ac data pairs are plotted to asses the ability of the discriminator to learn the conditionality automatically. The \ac losses remain high for the baseline and converge to $0$ for the proposed \ac cGAN. Figure~\ref{fig:dist} presented the histogram results for this experiment showed that the proposed approach better models conditionality. Subsequently the gradients provided to the generator are expected to be better directed. The gradient norm is plotted in Figure~\ref{fig:dynamics}.

\begin{table}[]   
    \centering
    \resizebox{0.8\textwidth}{!}{\begin{tabular}{|c||c|c|c c|}
         \hline
         Method & Resolution & FID & mIoU & Pixel accuracy (PA) \\
         \hline
         pix2PixHD & $256\times512$  & 66.7 & 56.9 & 92.8 \\
         \ac pix2pixHD &  $256\times512$ & \textbf{60.1}& \textbf{60.1} & \textbf{93.2}\\
         \hline
         SPADE & $256\times512$ & 65.5 & 60.2 & 93.1\\
         \ac SPADE & $256\times512$ & \textbf{59.9} & \textbf{61.5} & \textbf{93.7} \\
         \hline
         CC-FPSE & $256\times512$ & \textbf{52.4} & 61.8 & 92.8\\ 
         \ac CC-FPSE & $256\times512$ & 53.5 & \textbf{63.9} & \textbf{93.5} \\
         \hline
    \end{tabular}}
        \vspace{0.2cm}
    \caption{ \small	A comparison of different architectures trained from scratch with and without \ac augmentation. The networks with \ac achieves better results with a mean improvement of $\Delta mIoU= +2.3$, $\Delta PA= +0.56 $, and $\Delta FID= -3.8$.}
    \label{tab:high_resolution}
\end{table}

\subsection{"Single-label"-to-image}
The generality of the proposed \ac cGAN can also be demonstrated by showing that it also improves architectures other than image-to-image. An example of a different task is conditioning the generated image on a single input class-label as in~\cite{Brock2019LargeSG,Miyato2018cGANsWP,Odena2017ConditionalIS,Karras2020TrainingGA,Kang2020ContraGANCL}. This different architecture is of interest because many new methods for improving cGANs are often tested on this task. Unfortunately, these methods are mainly evaluated on the FID~\cite{Heusel2017GANsTB} and IS~\cite{Salimans2016ImprovedTF} scores. As stated earlier, these metrics measure the quality/diversity and they favor models that memorise the training set~\cite{gulrajani2020towards}. They have not been designed to evaluate conditionality and therefore not sufficient for the purpose of this paper. 
Despite that, these criteria are still important for evaluating the quality of GANs, however, an additional criterion is required for testing conditionality. 

Here a simple conditionality test is proposed specifically for "single label"-to-image generation tasks based on a pretrained Resnet-56~\cite{he2016deep} classifier trained on CIFAR-10~\cite{cifar}. BigGAN~\cite{Brock2019LargeSG} was selected as the baseline. Since BigGAN uses the Hinge-loss~\cite{lim2017geometric}, the \ac loss is adapted as follows:
\begin{align}
        \mathcal{L}_{D}=  -&\mathbb{E}_{\mbf{x} \sim p(\mbf{x}) , \mbf{y} \sim p(\mbf{y|x})}\big[min(0, -1 + D(\mbf{x},\mbf{y})]\big] -  \mathbb{E}_{\mbf{x}\sim p(\mbf{x})}\big[min(0,-1 - D(\mbf{x},G(\mbf{x}))]\big]  \notag  \\  
        -&\mathbb{E}_{\mbf{\tilde x} \sim p(\mbf{\tilde x}) , \mbf{y} \sim p(\mbf{y})}\big[min(0,-1-D(\mbf{\tilde x},\mbf{y}))\big] - \mathbb{E}_{\mbf{\tilde x}\sim p(\mbf{\tilde x}), \mbf{x} \sim p(  \mbf{x})}\big[min(0,-1 - D(\mbf{\tilde x},G(\mbf{x})))\big] \notag \\
        \mathcal{L}_{G}= - &\mathbb{E}_{\mbf{x}\sim p(\mbf{ x})} D(\mbf{x},G(\mbf{x}))
\label{eq:HingleLoss}
\end{align}
Both models are trained from scratch on CIFAR-10~\cite{cifar} dataset using the hyper-parameter specified in~\cite{Brock2019LargeSG}. The conditionality is tested by generating 10k images for each label(100k images in total) and calculating the accuracy. The results\footnote{The Pytorch IS and FID implementations were used for comparison}are shown in Table~\ref{tab:biggan_table}.
\begin{table}[h]   
    \centering
    \begin{tabular}{|c||c|c|c|c|}
         \hline
         Method & IS score & FID score & Acc  \\
         \hline
         BigGAN\cite{Brock2019LargeSG} & 8.26 $\pm$ 0.095   & 6.84  & 86.54 \\
         \hline
         \ac BigGAN & \textbf{8.40} $\pm$ 0.067 &   \textbf{6.28} & \textbf{92.04} \\
         \hline
    \end{tabular} 
        \vspace{0.2cm}
    \caption{\small A comparison of BigGAN~\cite{Brock2019LargeSG} with and without the \ac GAN. The network with \ac achieves significantly better results with an improvement of $\Delta Acc= +5.59$, $\Delta IS= +0.14 $, and $\Delta FID= -0.56$.
    }
    \label{tab:biggan_table}
\end{table}

The conditionality improved significantly over the baseline with $\Delta Acc=+5.59$ and the quality also improved with $\Delta FID= - 0.56\:$, $ \Delta{IS}=+0.14$. Similar to the observation made before \ac enforces the conditionality without compromising the quality. A failure mode of the lack of conditionality of the discriminator is class leakage : images from one class contain properties of another. While is it not easy to define a proper metric for such failure mode, it is shown that using the \ac loss the classification was improved and therefore the generation is better constrained and does not mix class properties. This result shows that \ac GAN also improves on a different SOTA task and confirms again that conditionality is an overlooked factor in current SOTA metrics.

\subsection{Monocular Depth prediction}

Monocular depth prediction is an ill-posed problem as an infinite number of 3D scenes can be projected onto the same 2D scene. A good model for predicting image geometry should exploit visual clues such as object sizes, lighting, shadows, object localization, perspective and texture information. Depth prediction is an appropriate task to evaluate the conditional relationship between $\mbf{x}$ and $\mbf{y}$. The RMSE $\log$ error is a strong metric to evaluate the conditional information since it takes into account the fidelity of the prediction with respect to the input. Similar to~\cite{eigen2014depth} other metrics are also reported for completeness. The model is trained on the NYU Depth V2 Dataset~\cite{Silberman:ECCV12} to predict depth from monocular 2D-RGB images only. The official train/validation split of $795$ pairs is used for training and $694$ pairs are used for validation. Similarly to low resolution label-to-image translation, the dataset images are resized to have a resolution of $256\times256$. The experiment is repeated 6 times without fixing the seed and the mean and standard deviation are reported.
\begin{figure*}
\centering   
\subfigure{\includegraphics[width=0.3\textwidth]{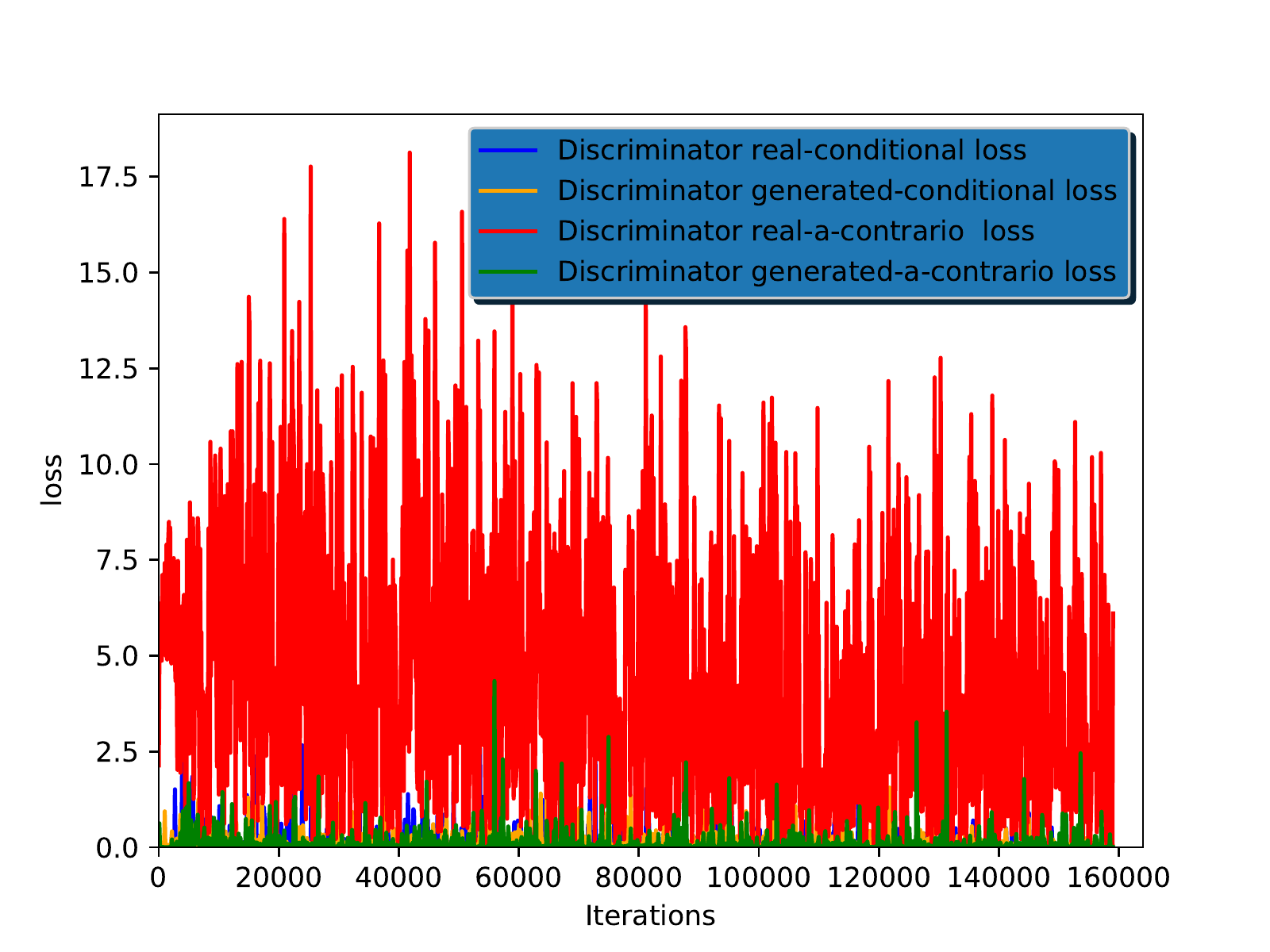}}
\subfigure{\includegraphics[width=0.3\textwidth]{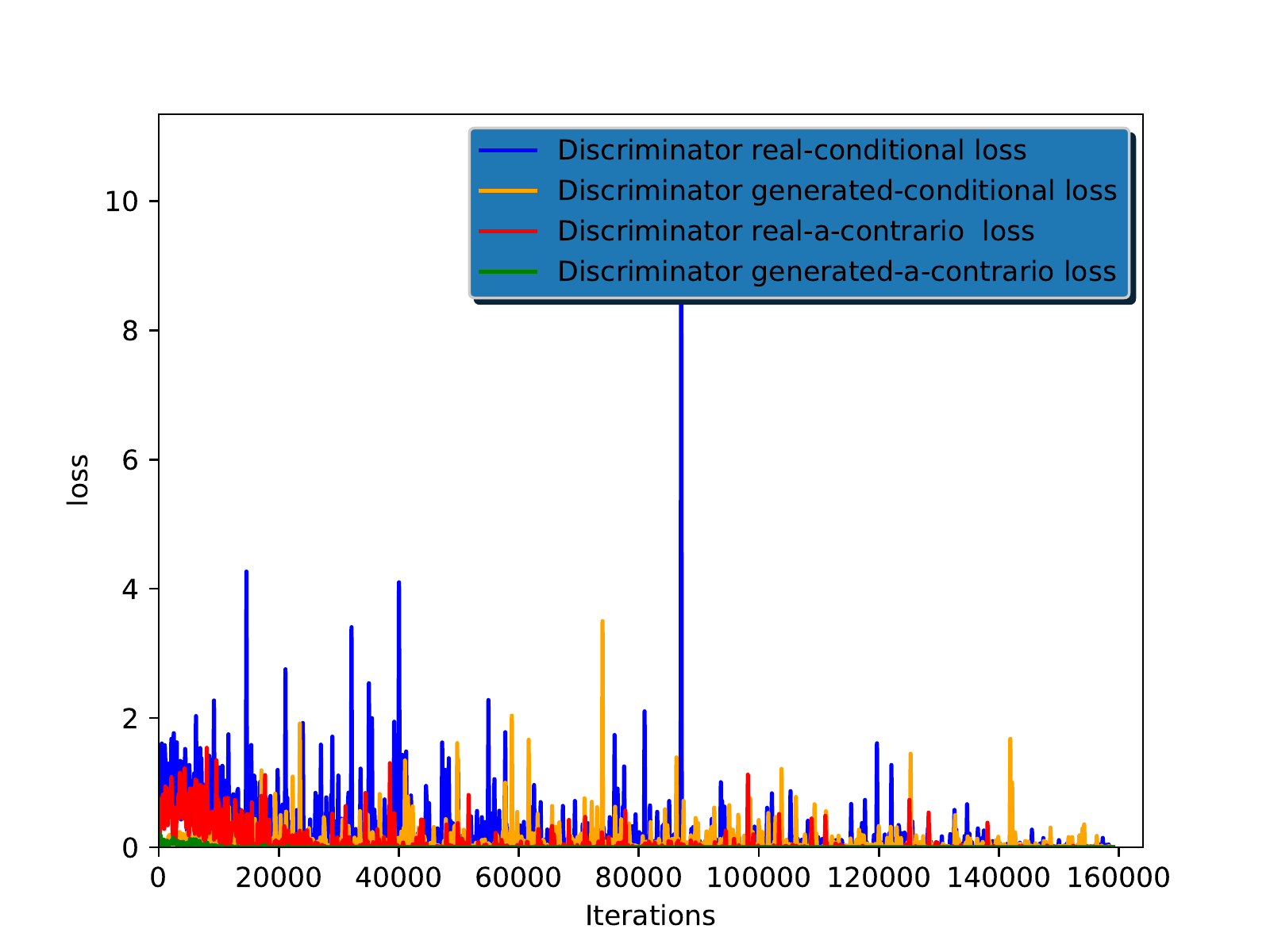}} 
\subfigure{\includegraphics[width=0.3\textwidth]{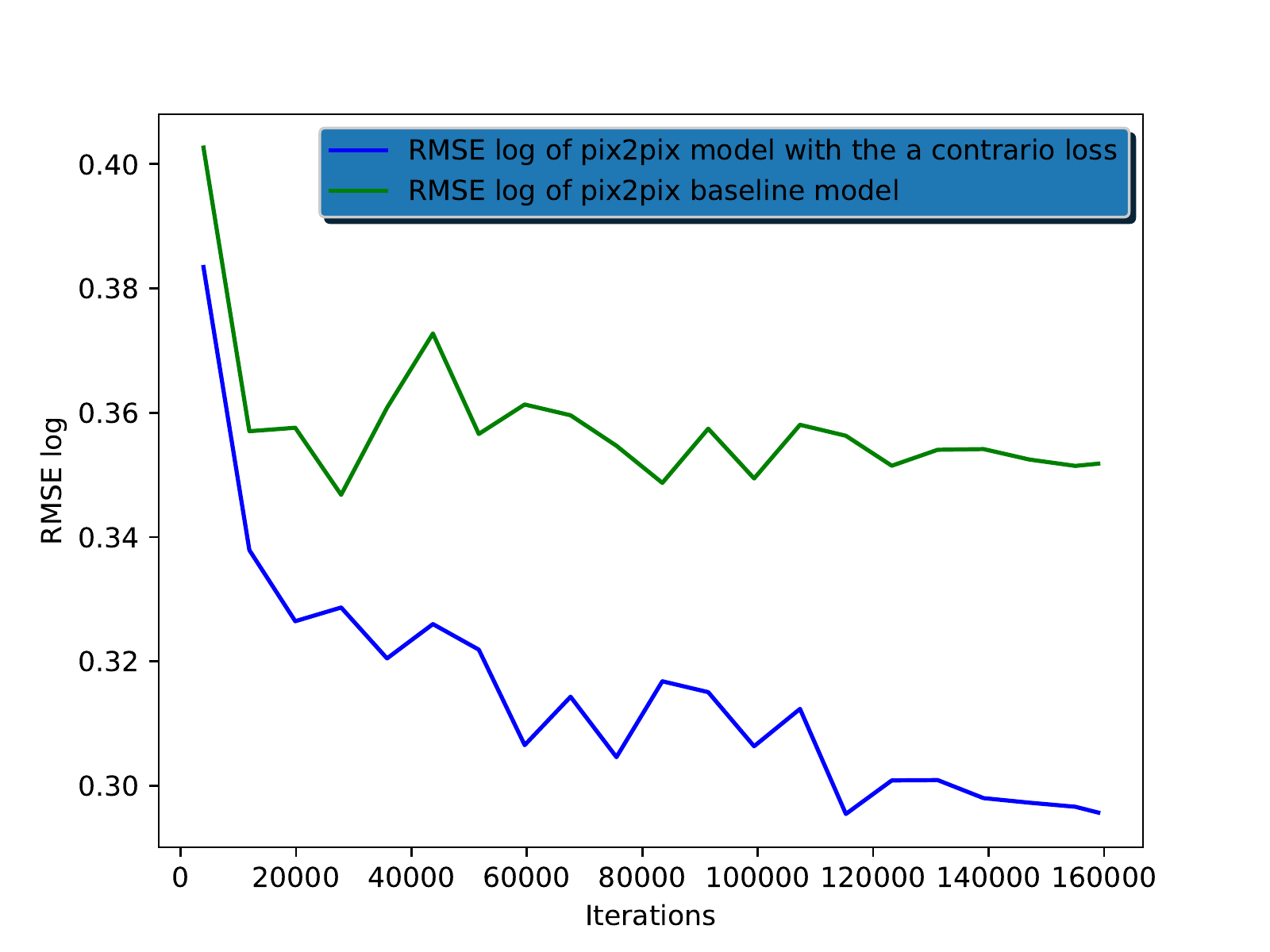}} 
    \caption{\small	Comparison of the proposed approach on the NYU Depth V2 training set. (a) The loss function for each set of data-pairing for the baseline cGAN method. (b) The loss function for each set of data-pairing for the \ac cGAN method. (c) The evolution of the RMSE $\log$ for both methods, performed on the validation dataset. It can be seen in (b) compared to (a) that the \ac loss converges to 0 rapidly for the proposed approach. In (c) the proposed approach is much more efficient and converges much faster and with much higher accuracy.}
    \label{fig:depth}
\end{figure*}

Figure~\ref{fig:depth}(c) shows the RMSE $\log$ performances of the validation set over the training iterations for one experiment. Table~\ref{tab:Depth_stat} shows the comparison of the two models across different metrics. Clearly the \ac cGAN reaches a better performance with log RMSE $0.3036$ versus $0.3520$ for the baseline (the mean is reported here). The evaluation of mode collapse using NDB~\cite{richardson2018gans} of the two networks is given in Section~\ref{sec:mode-collapse}. The qualitative results are shown in Figure~\ref{fig:depth_qualitative}. Moreover, Figure~\ref{fig:depth}(a) and Figure~\ref{fig:depth}(b) compare the losses of the discriminator for the different data pairings when using the baseline model and the proposed \ac model. Similar to Label-To-Image translation, the baseline results show high loss for real \textit{a contrario}. 

The discriminator is optimized only to distinguish real and generated samples its decision boundary is independent off conditional variable. The baseline cGAN architecture will not penalize the generation of outputs belonging to the target domain but that do not correspond to the input. Not only does this leave the generator with a larger search space ( the generator is less efficient), but it can allow mode collapse whereby the generator always produces the same output. The \ac loss explicitly avoids this by penalizing unconditional generation. The \ac losses of the \ac cGAN can be observed in Figure~\ref{fig:depth}(a) and Figure~\ref{fig:depth}(b) to converge rapidly to 0. This indicates the search space of the generator is explicitly restricted to only conditional samples early in the training. 
\linespread{0.6}
\begin{table}[]   
    \centering
    \begin{tabular}{|c||c|c|c|c|}
         \hline
         Method &  RMSE $\log$ & si$\log$ & $\log_{10}$ & abs rel \\
         \hline
         baseline & 0.3520$\pm$ 0.0016 & 28.54$\pm$ 0.1932  & 0.1247 $\pm$ 0.0005 & 0.3318$\pm$ 0.0026 \\
         \hline
         \ac & \textbf{0.3036}$\pm$ 0.0055  & \textbf{23.51}$\pm$ 0.1932 & \textbf{0.1079} $\pm$ 0.0021 & \textbf{0.2868} $\pm$ 0.0093\\
         \hline
    \end{tabular}
    \vspace{0.2cm}
    \caption{\small	 Monocular Depth prediction experiments were repeated on the baseline and \ac cGANs 6 times with different seeds. The mean and standard deviation are reported for each metric. The results shows that the \ac cGAN outperforms the baseline on the depth metric~\cite{eigen2014depth}.
    }
    \label{tab:Depth_stat}
    
\end{table}

\subsection{Image-to-label segmentation}
Image-to-label is a simpler task compared to depth prediction and label-to-image prediction as the goal of the generator is to transfer from a high-dimensional space to a lower-dimensional space. Furthermore, the evaluation is simpler since the image mask does not have multiple solutions and it is not necessary to use an external pre-trained segmentation network for comparison as in the case of label-to-image translation. It is worth mentioning that pix2pix is trained to output 19 classes as a segmentation network and is not trained as an image-to-image network as it is often done in cGAN architectures. FCN~\cite{long2015fully} trained on~\cite{Isola2017ImagetoImageTW} obtains $21.0$ mIoU. The performances are shown in Table~\ref{tab:seg_table}. As expected the training was unstable (refer to Figure~\ref{fig:city}). However, the \ac cGAN shows superior mIoU performance with $19.23$ versus $15.97$ for the baseline model. Figure~\ref{fig:seg_qualitative} shows the qualitative results of the both models. It can be observed that the model baseline has invented labels that are not specified by the input. Training with \ac helps the discriminator to model conditionality. Thus, the generator search space is restricted to only conditional space. The generator is penalized for conditionality even if the generation is realistic.   
\begin{table}[]   
    \centering
    \begin{tabular}{|c||c|c|c|c|}
         \hline
         Method & Pixel accuracy (PA) & Mean Acc & FreqW Acc  & mIoU  \\
         \hline
         Baseline & 66.12 & 23.31  & 53.64  & 15.97 \\
         \hline
         \ac & \textbf{72.93} &  \textbf{26.87} & \textbf{60.40} & \textbf{19.23} \\
         \hline
    \end{tabular} 
        \vspace{0.2cm}
    \caption{\small Comparison on the Cityscapes dataset validation set. The proposed method consistently obtains more accurate results and finishes with a largely different score at the end of training with mIoU of $19.23$ versus for the baseline $15.97$.}
    \label{tab:seg_table}
\end{table}

\section{Conclusion}
 
This paper has proposed two important contributions for conditional Generative Adversarial Networks (cGANs). The first main contribution is a probabilistic analysis of the discriminator to show that it is not explicitly conditional. The second contribution is a new method, called \textbf{\ac~cGAN}, that explicitly models conditionality for both parts of the adversarial architecture via a novel \ac loss that involves training the discriminator to learn unconditional (adverse) examples. This leads to a novel type of data augmentation approach for GANs (\ac learning) which allows to restrict the search space of the generator to conditional outputs using adverse examples. Extensive experimentation has shown significant improvements across several tasks, datasets and architectures. 

Although \ac has shown significant improvement for image domain addressing other modalities such as audio and text is to be tested in the future work. It will also be of interest to explore multiple discriminator output classes for the four pairings considered in this paper.

\section{Acknowledgements}
This work was performed using HPC resources from GENCI-IDRIS (Grant 2021-011011931). The authors would like to acknowledge the Association Nationale Recherche Technologie (ANRT) for CIFRE funding (n°2019/1649).

\linespread{0.9}
\bibliography{egbib}
\newpage
\appendix
\counterwithin{figure}{section}

\section{Supplementary material}

Supplementary material is presented here as follows, first Section~\ref{sec:histeval} provides more extensive results showing that the histogram evaluation method successfully captures and shows the various statistical modalities related to the various pairings in the dataset. Section~\ref{sec:conditionanalysis}, provides a further analysis on the conditionality of cGAN and the proposed method via carefully selected boundary cases which show the failure of the classic discriminator to learn conditionality. Section~\ref{sec:mode-collapse} provides an additional evaluation of mode collapse for the depth prediction model. Section~\ref{sec:loss_extra} looks into the choice of weighting the different parts of the proposed loss function. Details are provided for reproducibility in Section~\ref{sec:implementation}. Finally, an analysis of the training procedure is provided in Section~\ref{sec:dynamics} to show that the training procedures did not encounter any degenerate situations. 
 
\subsection{Histogram evaluation criteria}
\label{sec:histeval}

\begin{figure*}[]
    \centering
    \subfigure{\includegraphics[width=\textwidth]{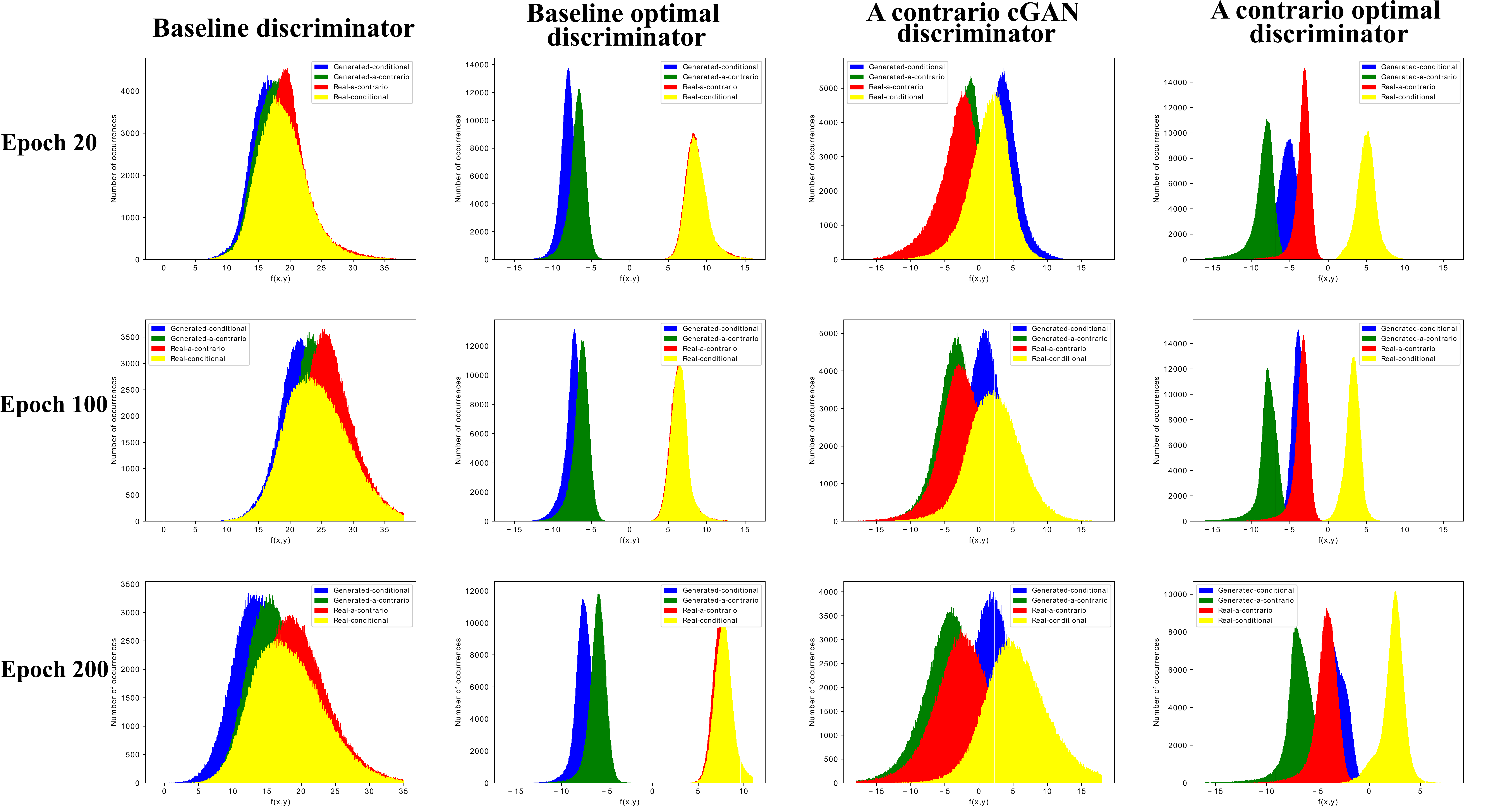}}\\
    \subfigure{\includegraphics[width=\textwidth]{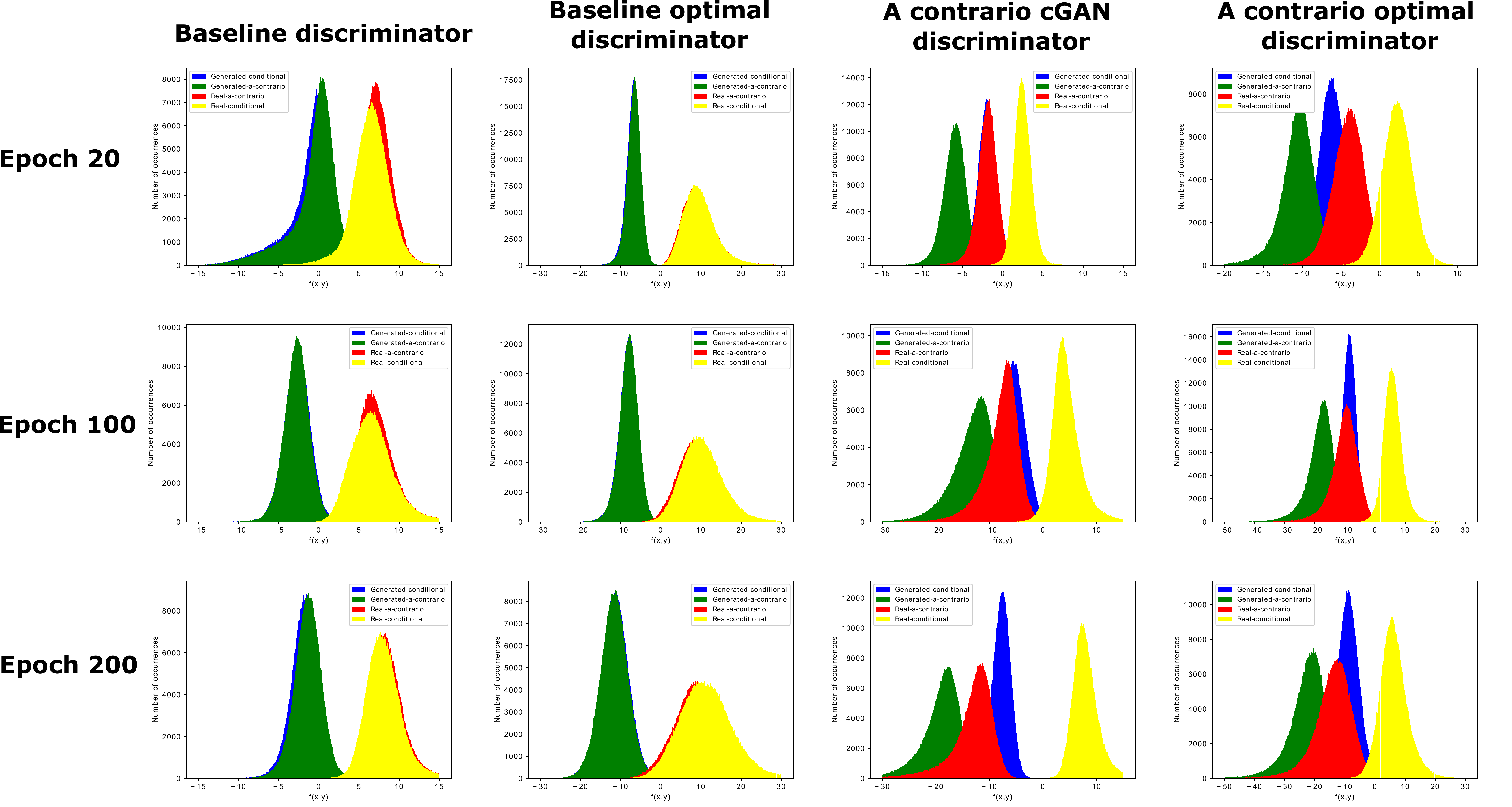}}
    \caption{Yellow and Blue are the classic data pairings being that of real-conditional and generated-conditional respectively. The proposed \ac data pairings are Red and Green for real-\act and generated-\act respectively. The last convolution layer $f(\mbf{x},\mbf{y})$ shows positive values for "true" and negative values for "false" since the Sigmoid activation was used for training. The histogram evaluation is performed with dropout and batch normalization during training. The top three rows show Cityscape label-to-image histograms. The bottom three rows show the NYU Depth monocular depth prediction histograms. The \ac cGAN optimal discriminator correctly classifies each data pairing while the PatchGAN optimal discriminator fails by classifying real-a-contrario as true (greater than zero).}
    \label{fig:histograms-appendix}
\end{figure*}

The discriminator encodes the high dimensional space of input pairs into a lower dimensional latent space. Visualizing the empirical distribution on a high dimensional space is infeasible, however, since the encoded latent space learnt by the discriminator is compact, it is possible to visualize the empirical distribution by observing the latent space. This provides a direct insight into the discriminator performance and insight into the errors fed back for training the generator. For the purpose of this paper, a histogram is plotted in order to demonstrate the capacity of the discriminator to correctly classify underlying data distributions (see Figure~\ref{fig:histograms-appendix}) as an example. A good cGAN discriminator should classify real-conditional as true, and all the three remaining pairs as fake even if the variables of the pairs are sampled from real data distribution (the case of \ac real). Further analysis and results are provided to show the performance of the evaluation using the proposed histogram approach. As a reminder, 4 sets of data pairings are created as formalised in Section~\ref{sec:ccgan}. The output response of the discriminator for each of these pairings is then plotted as a histogram with a different color. It can be noted that in the original GAN paper~\cite{goodfellow2014generative}, the authors provide the intuition behind the underlying probability distributions. With the histogram test approach shown here, these distributions can be clearly observed.

The work presented in~\cite{goodfellow2014generative} showed that the optimal discriminator converges to: 
\begin{align}
    D^* = \frac{p(\mbf{y}|\mbf{x})}{p(\mbf{y}|\mbf{x}) + p_{g}(\mbf{y}|\mbf{x})} \label{eq:optimaldiscriminato_expression}
\end{align}
Therefore considering the GAN cost function in Eq~\ref{eq:cost} with the definition in Eq~\ref{eq:classic_loss} the cost function becomes: 
\begin{align}
    V(G,D^*) = \min_{G} -log(4) + 2 D_{js}(p_{g} || p) \label{eq:shannon_janson_div}
\end{align} 
$p(\mbf{x},\mbf{y}|\mbf{x})$ is the real-conditional pair distribution while $p_{g}(\mbf{x},\mbf{y}|\mbf{x})$ is the generated-conditional pair. Therefore, Eq~\ref{eq:cost} corresponds to minimizing Jensen Shannon divergence between the probability distribution defined by the generator $p_g(\mbf{x}, \mbf{y}|\mbf{x})$ and the real probability distribution $p(\mbf{x}, \mbf{y}|\mbf{x})$. Therefore, for an optimal discriminator, the optimal generator is defined when $p_g(\mbf{x}, \mbf{y}|\mbf{x})$ matches the real probability distribution $p(\mbf{x}, \mbf{y}|\mbf{x})$. If the optimal discriminator fails to model conditionality, the generator may not be able to match the real probability distribution $p(\mbf{x}, \mbf{y}|\mbf{x})$. That is why GAN models for semantic image synthesis suffer from poor image quality when trained with only adversarial supervision \cite{Sushko2020YouON} and consider additional loss terms such as perceptual loss \cite{Johnson2016PerceptualLF}, L1\cite{Isola2017ImagetoImageTW} or feature-matching \cite{Salimans2016ImprovedTF}. The segmentation-based discriminator proposed in \cite{Sushko2020YouON} can be considered as a strong conditional discriminator by construction. The purpose of that paper was not explicitly enforcing conditionality, however, following the results presented in this paper, it could be concluded that they reach the state of the art by their task specific error term which induces strong conditionality on the discriminator.

Figure~\ref{fig:histograms-appendix} shows the histograms of label-to-image and monocular depth prediction for epochs $20$, $100$ and $200$. Several observations can be made:
\begin{itemize}
    \item For the two tasks and for all evaluated epochs, the \ac cGAN optimal discriminator correctly classifies each data pairing. The classical cGAN optimal discriminator fails by classifying real-\act as true (greater than zero). 
    \item For \ac cGAN, during training, the conditional-generated samples are shifting towards towards the positive side of the function while training through the various epochs. This shows that the generator is correctly learning to  minimize the divergence between the generated-conditional and real-conditional sets of data. 
    
    \item Unlike classification metrics like accuracy, this histogram analysis provides more insight. For instance, even though the generated-conditional, generated-\act and real-\act are being classified as fake, three distributions are clearly distinguishable. There are therefore three different modalities for fake classification. There is only a single modality for the "true" classification which is clearly for real-conditional in yellow. The column showing the baseline optimal discriminator can be observed to model real-conditional and real-\act with approximately the same distribution (i.e. only 1 mode is visible for both of these pairings). This indicates that the discriminator is invariant to the conditionality.
\end{itemize}
\begin{figure}
\centering    
    \includegraphics[width=0.7\textwidth]{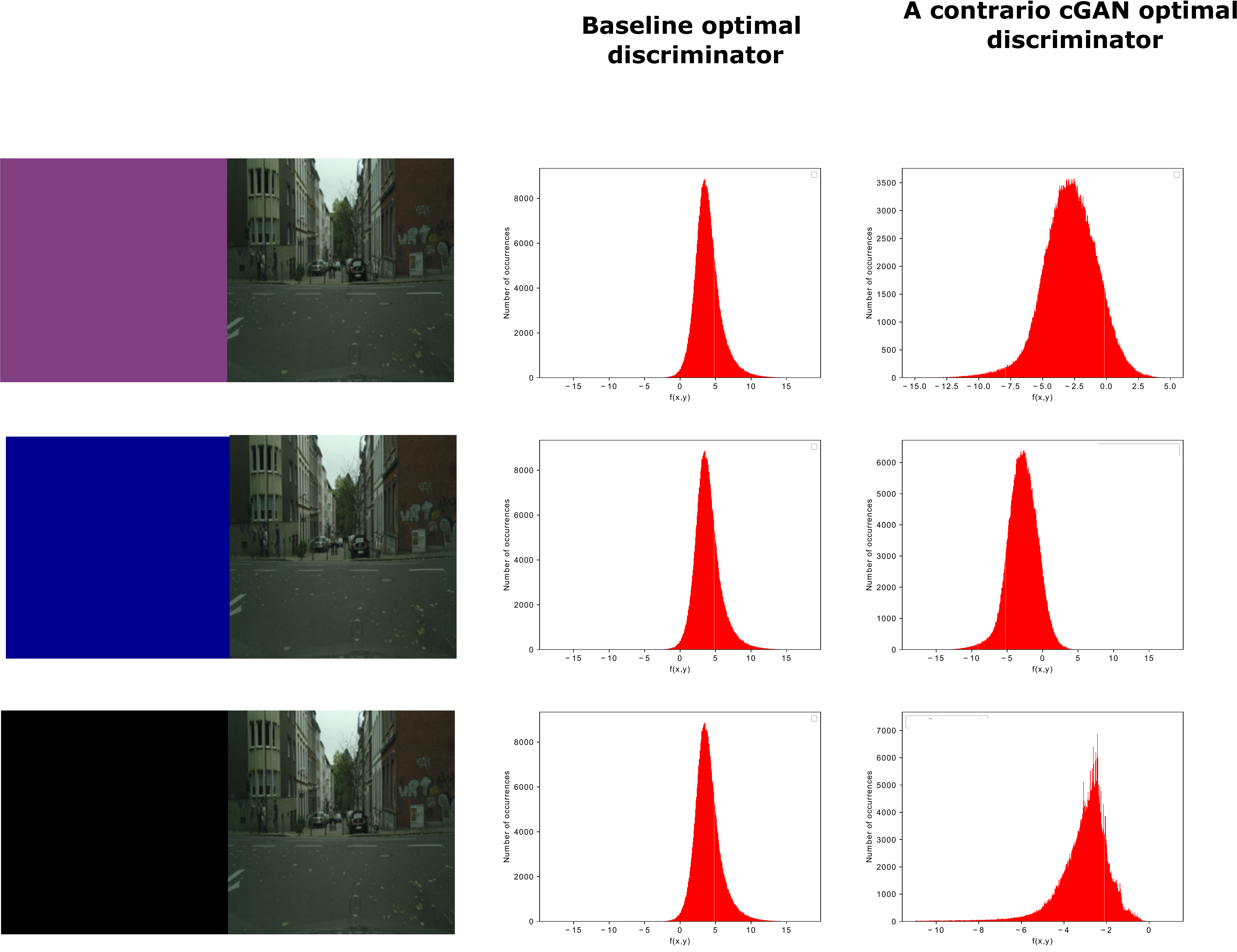}
    \caption{Extreme cases are provided to evaluate the two discriminators. The first row represents the "all-road" label for all the 500 validation images. The second row represents the "all-car" label. The third row represents the "no-object" label. It can be seen clearly that the PatchGAN discriminator fails to classify these pairs as fake while \ac cGAN succeeds to classify them correctly.}
    \label{fig:conditionality}  
\end{figure}

\subsection{Conditionality analysis}
\label{sec:conditionanalysis}
In addition to Figure~\ref{fig:pipeline} in the paper, Figure~\ref{fig:conditionality} is provided here to show various failure cases for the classic cGAN approach. The test was carried out on the Cityscapes dataset label-to-image trained as mentioned in Section~\ref{sec:experiments}. The histogram is plotted on the $500$ Cityscapes where $\mbf{y} \sim p(\mbf{y}|\mbf{x})$ . The experiment was performed using the optimal discriminator for both models. 

Extreme cases are chosen to assess the PatchGAN discriminator. Providing an "all-road", "all-car", "no-object" label for each pixel paired along the set of real images as an input pair. It can be seen clearly that PatchGAN discriminator classifies these pairs as "true". The \ac cGAN successfully classifies these pairs as "fake". This suggests that the PatchGAN focuses only on $\mbf{y}$ instead of looking at the pair $(\mbf{x},\mbf{y})$. 

Most PatchGAN-based methods do not pay careful attention to data pairing when training the discriminator, and subsequently the same conditional input image is reused for both real and generated input pairs in each mini-batch. More precisely, the conditional variable $\mbf{x}$ is always the same in each condional pairing and only $\mbf{y}$ changes ($\mbf{y}\sim p(\mbf{y}|\mbf{x})$ or $\mbf{y}\sim p_g(\mbf{y}|\mbf{x})$). Subsequently the discriminator network learns to ignore the conditional input and the predicted true/fake label is only determined from the variable $\mbf{y}$. During experimentation we tried to resolve this issue by not allowing the same $\mbf{x}$ to appear in both generated and real  input pairings within a mini-batch. This test yielded the same results which suggest that ignoring the conditional variable is a fundamental problem of the classic PatchGan architecture. As mentioned earlier, other architecture were also tested for conditionality and the result was the same. This suggest that that the problem is not specific to PatchGAN but generalises to cGAN architectures.

\subsection{Mode collapse analysis}
\label{sec:mode-collapse}
Mode collapse is the setting in which the generator learns to map several different inputs to the same output. A collapsing model is by construction unconditional. Only a few measures have been designed to explicitly evaluate this issue~\cite{richardson2018gans,wang2003multiscale,arora2018gans}. MS-SSIM~\cite{wang2004image,wang2003multiscale} measures a multi-scale structural similarity index and birthday paradox~\cite{arora2018gans} concerns the probability that, in a set of $n$ randomly chosen outputs, some pair of them will be duplicates. Another approach, NDB~\cite{richardson2018gans}, presents a simple method to evaluate generative models based on relative proportions of samples that fall into predetermined bins. 

\begin{figure}
    \centering
    \includegraphics[width=0.7\textwidth]{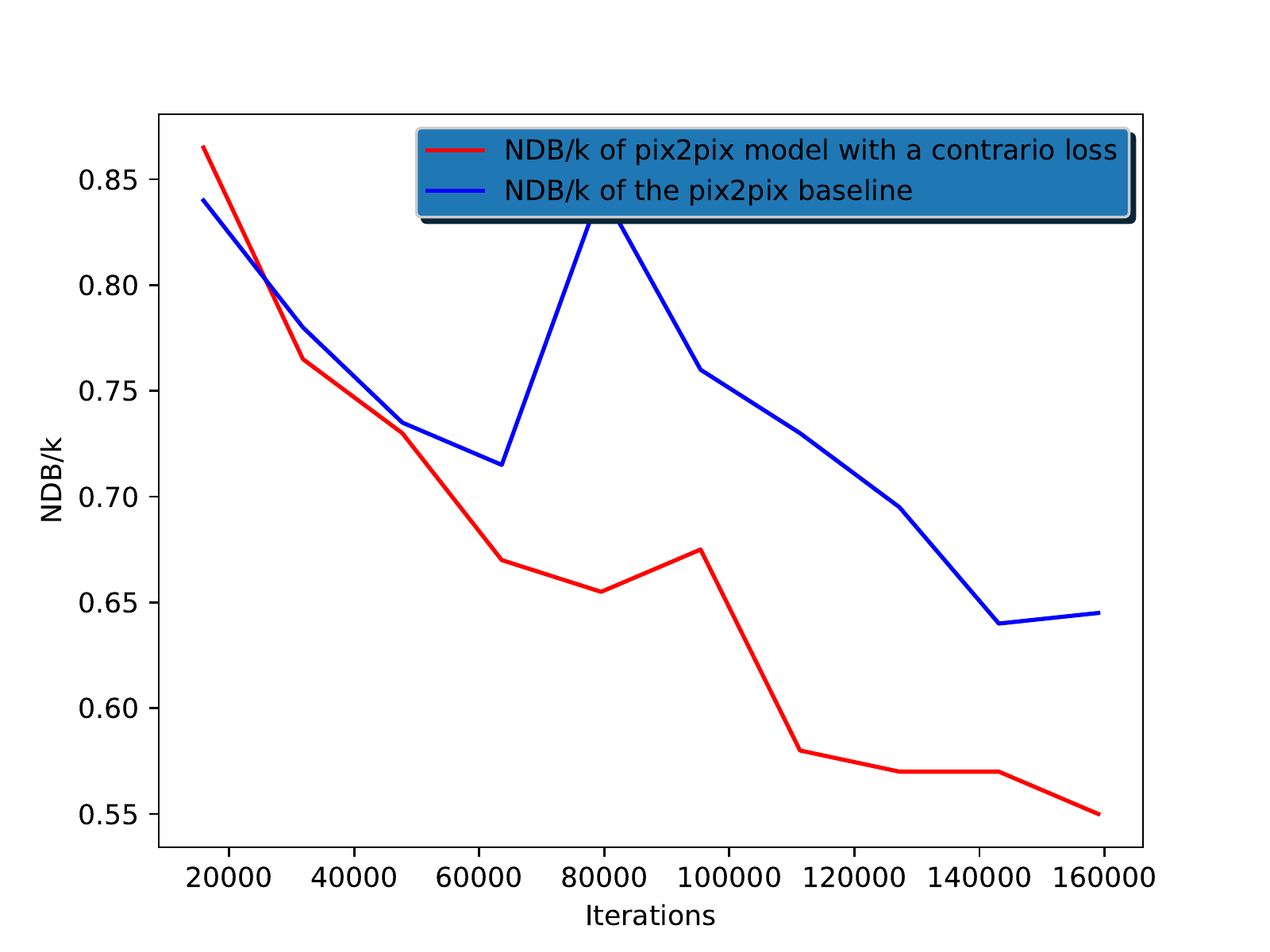}
    \caption{An analysis of mode collapse using the NDB criteria (lower values are better) throughout training on the NYU depthV2 dataset. It can be concluded from this evaluation that the proposed approach is much better at avoiding mode collapse due to the restricted search space of the generator.}
    \label{fig:collapse}
\end{figure}

\begin{figure*}
    \centering
    \includegraphics[width=0.8\textwidth]{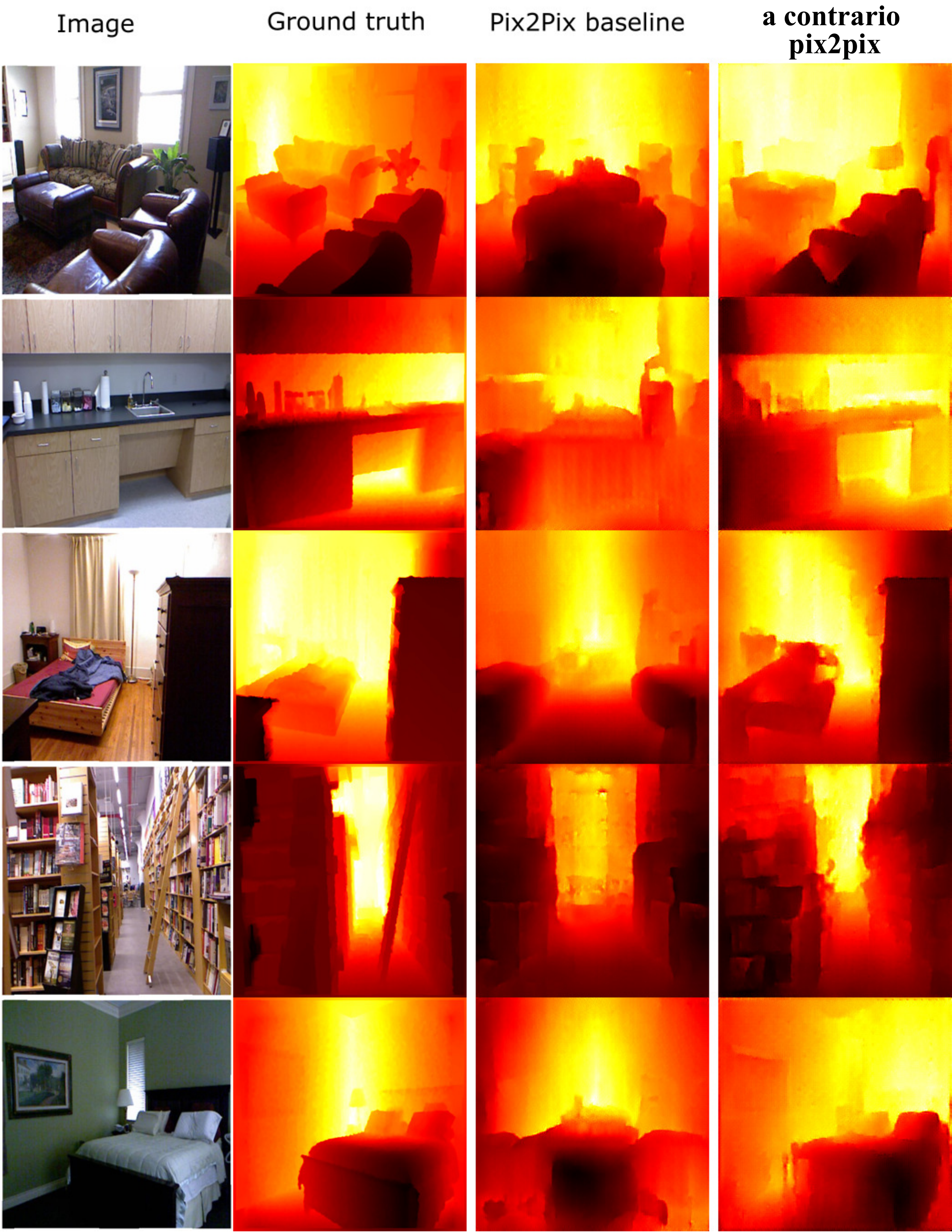}
    \caption{Qualitative results for depth prediction. The \ac cGAN shows better performance and more consistent prediction with respect to the input. The first row shows a case of mode collapse for the baseline as it ignores completely the input.}
    \label{fig:depth_qualitative}
\end{figure*}

The analysis provided in this section is an extension of the experiments done on depth prediction. Figure~\ref{fig:collapse} shows the evolution of the NDB measure over training iterations using the NDB score (the less, the better) for both pix2pix baseline and \ac cGAN models trained on the NYU Depth V2 training set~\cite{Silberman:ECCV12}. Out of the 12 trained models, the best model (in terms of RMSE $\log$) is chosen for the evaluation. For clustering and evaluating NDB, non overlapping patches of $64\times64$ are considered. At the end of the training the NDB/k ($k=100$) of the \ac cGAN is $0.550$ while the baseline achieves only $0.645$. This indicates that \ac model \textbf{generalizes better}. This is also observed qualitatively in Figure~\ref{fig:depth_qualitative}. Training with the counter examples helps the discriminator to model conditionality. Thus, the generator search space is restricted to only conditional space. The generator is penalized for non-conditionality even if the generation is realistic.

\subsection{Loss function analysis}
\begin{figure}
\centering    
    \includegraphics[width=0.7\textwidth]{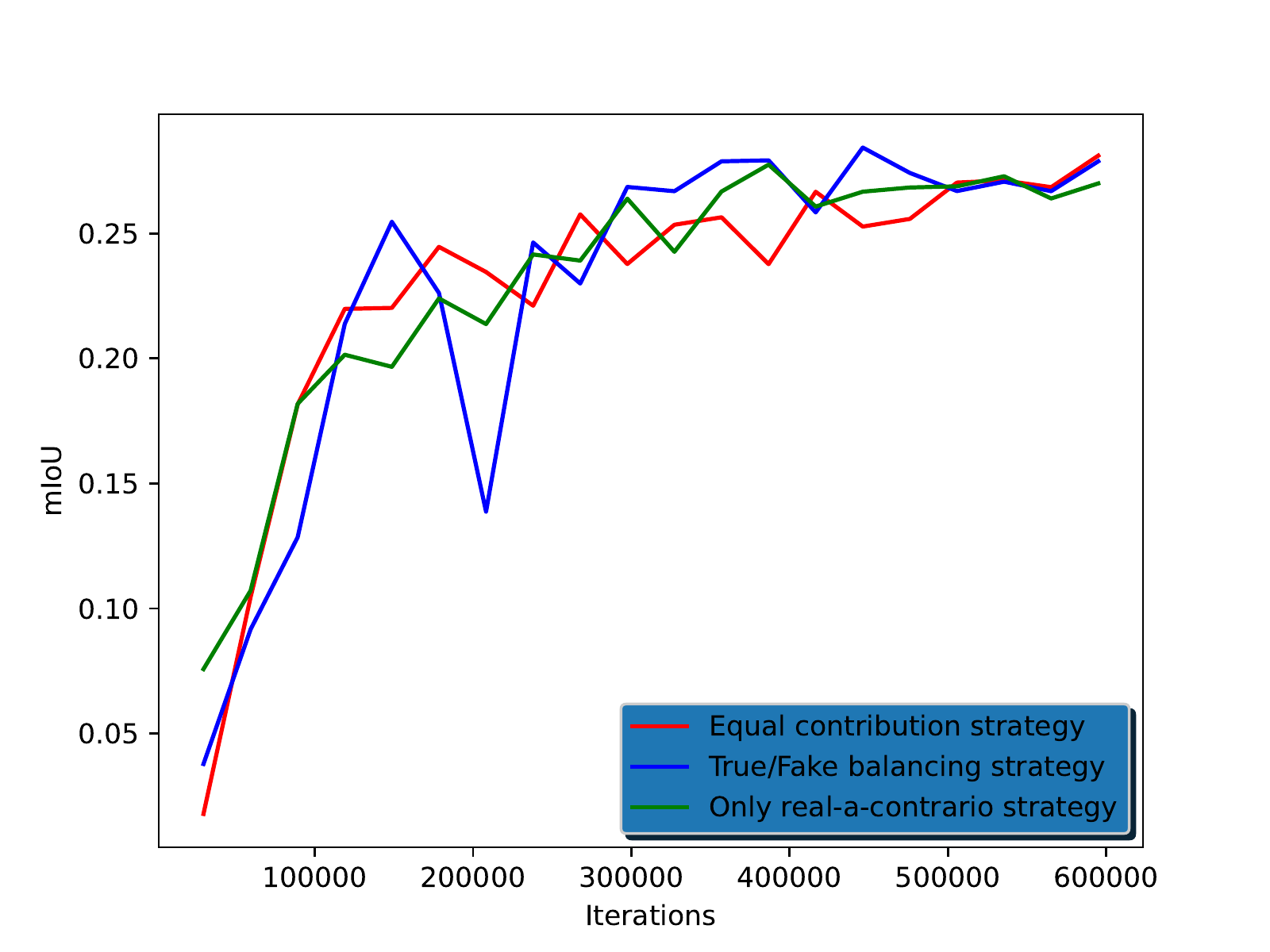}
    \caption{The mIoU evaluation for different choice of $\lambda_i$. The strategy 1 of giving equal contribution yield the best results. However, there is no major difference on the convergence or the performances at epoch 200 between the different strategies}
    \label{fig:lambda_choice}  
\end{figure}

An ablation study on Eq~\ref{eq:overall_loss} was performed. Each term that contributes to the adversarial loss is weighted by $\lambda_i$. Eq~\ref{eq:overall_loss} becomes: 
\begin{align}
    \mathcal{L}_{adv}= \min_{G} &\max_{D} \:  \Big[\lambda_1\mathbb{E}_{\mbf{x} \sim p(\mbf{x}) ,\mbf{y} \sim p(\mbf{y|x})}\big[log(D(\mbf{x},\mbf{y})]\big] + \lambda_2\mathbb{E}_{\mbf{x}\sim p(\mbf{x})}\big[log[1 - D(\mbf{x},G(\mbf{x}))]\big] \Big] +  \notag \\
    &\max_{D} \:\Big[ \lambda_3\mathbb{E}_{\mbf{\tilde x} \sim p(\mbf{\tilde x}) ,\mbf{y}\sim p(\mbf{y})}\big[log(1 - D(\mbf{\tilde x},\mbf{y}))\big]  + 
        \lambda_4\mathbb{E}_{\mbf{\tilde x} \sim p(\mbf{\tilde x}) ,\mbf{x} \sim p(\mbf{x})}\big[log(1 - D(\mbf{\tilde x},G(\mbf{x})))\big] \Big]
\end{align}
\label{sec:loss_extra}
Three strategies were considered for the weighting. The models were trained on the Cityscapes label-to-image dataset with the same settings described earlier (Section~\ref{sec:lb-to-im}). Figure~\ref{fig:lambda_choice} shows the mIoU for different \ac cGAN models trained with different choices  for $\lambda_i$. 
\begin{itemize}
\item \textbf{Strategy 1:} Equal contribution for each term : $\lambda_1 = \lambda_2 = \lambda_3 = \lambda_4$.

\item \textbf{Strategy 2:} Balancing the "fake" and "true" contributions. Since there are 3 data pairings classified as fake and only 1 real pair as true, equal balancing of true/fake gives: $\lambda_1 = 1 , \lambda_2 = \lambda_3 = \lambda_4 = 0.33$

\item \textbf{Strategy 3:} Testing the significance of both \ac error terms for fake and real images. In this case only 3 terms with real-a-contrario is tested : $\lambda_1 = \lambda_2 = \lambda_3 = 0.5, \lambda_4 = 0$.

\end{itemize}
In this simple test, Strategy 1 gives the best results. Strategy 2 seems less stable. Strategy 3 succeeds to learn conditionality, however, it may not capture conditionality for generated images during training. Each of these strategies succeed to model conditionality, however, Strategy 1 converges faster and yields a better final result in terms of mIOU. 

\subsection{Reproducibility}
\label{sec:implementation}

Various experiments were performed using different datasets and input-output modalities. Some extra detail is provided here for reproducibility purposes. In all the experiments using the pix2pix baseline, random jitter was applied by resizing the $256\times 256$ input images to $286\times 286$ and then randomly cropping back to size $256\times 256$. All networks were trained from scratch. Weights were initialized from a Gaussian distribution with mean $0$ and standard deviation $0.02$. The Adam optimizer was used with a learning rate of $0.0002$, and momentum parameters $\beta_1 = 0.5$, $\beta_2 = 0.999$. A linear decay is applied starting from epoch $100$, reaching $0$ at epoch $200$. Dropout is used during training. As in the original implementation~\cite{Isola2017ImagetoImageTW}, the discriminator is a PatchGan with a receptive field of $70\times 70$. Similarly pix2pixHD~\cite{Wang2018HighResolutionIS}, SPADE~\cite{Park2019SemanticIS} and CC-FPSE~\cite{Liu2019LearningTP} were trained with the same hyper-parameters as mentioned is their respective papers.  
For label-to-image, a U-Net256 with skip connections was used for the generator. A U-Net with $9$ ResNet blocks was used for depth prediction, the last channel is $1$ instead of $3$ and the activation of the last convolution layer generator is \textit{Relu} instead of \textit{Tanh}. 

For the image-to-label task, a U-Net256 with skip connections was used for the generator but the output channel size was chosen to be $19$ instead of $3$ for segmentation of $19$ classes. The activation of the last convolution layer of the generator was changed to a softmax to predict class probability for segmentation purposes.

\subsection{Training details}
\label{sec:dynamics}
\begin{figure}
    \centering
    \subfigure[]{\includegraphics[width=0.48\textwidth]{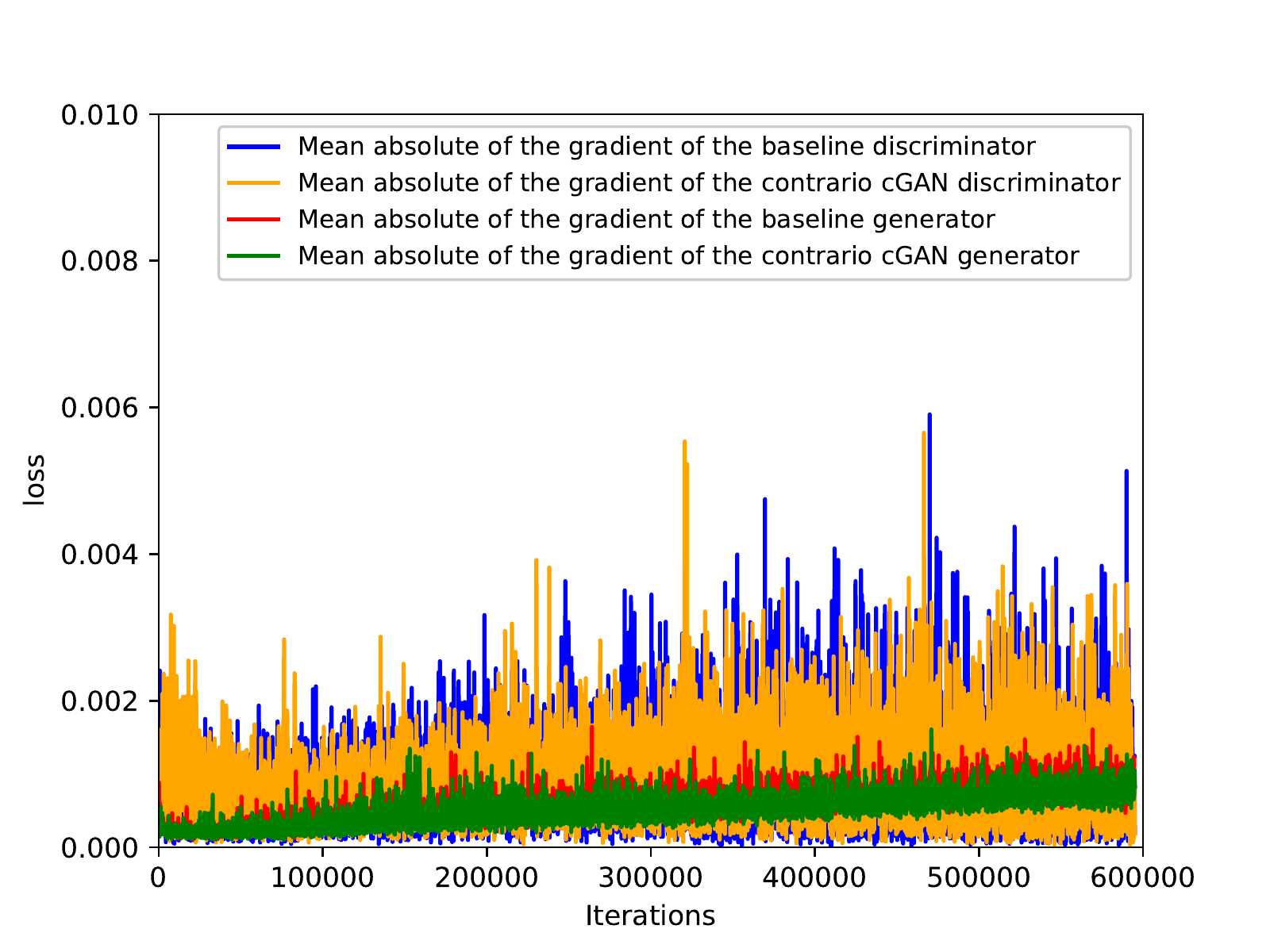}}
    \subfigure[]{\includegraphics[width=0.48\textwidth]{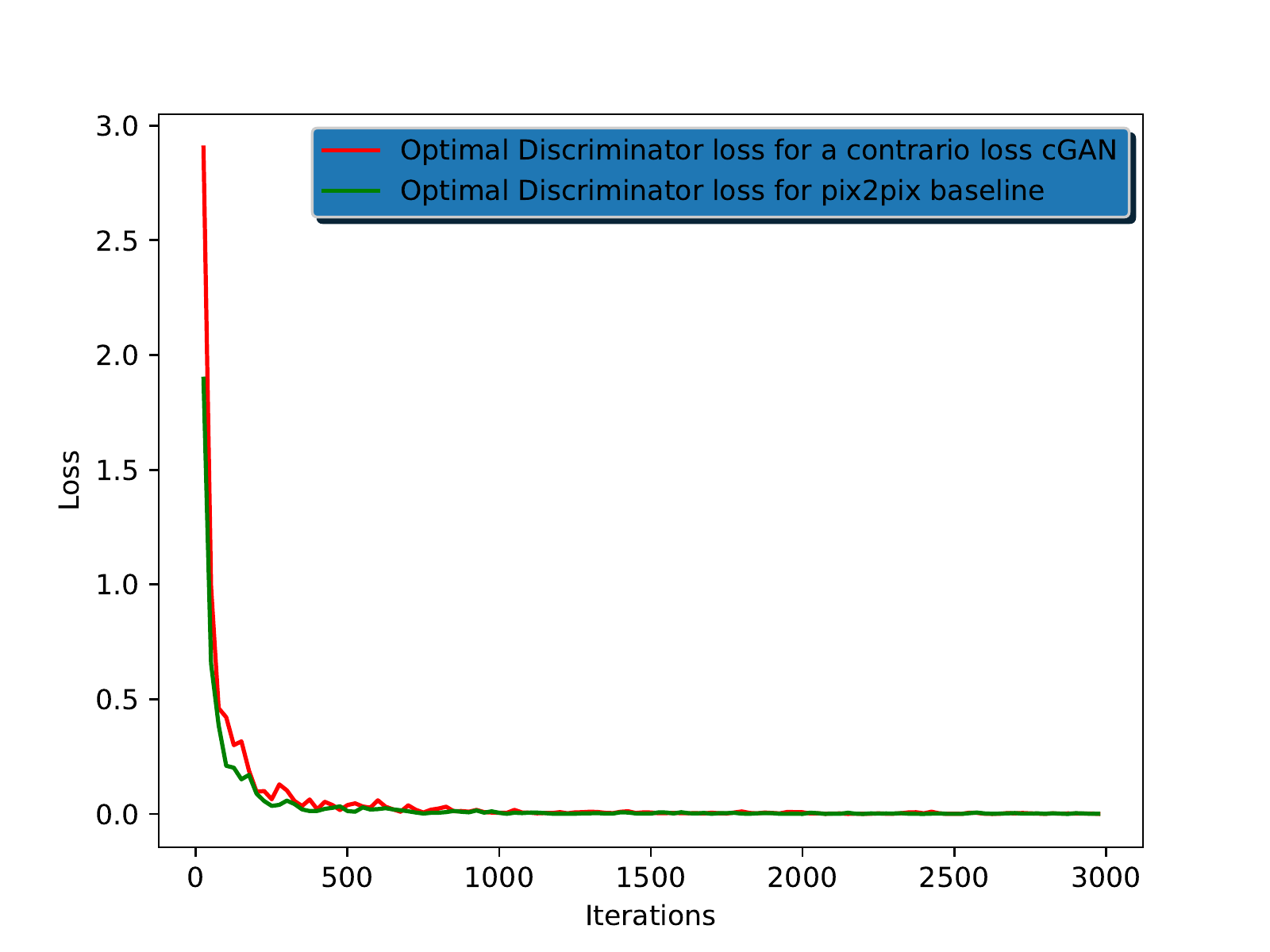}} 
    \caption{(a) The mean absolute value of the gradients of the generator and discriminator for both baseline and \ac cGAN models trained on Cityscapes\cite{Cordts2016Cityscapes}. The gradient is stable and it is neither vanishing nor exploding. (b) The loss function of the optimal discriminators when the generator is fixed. Both losses converge rapidly to $0$.}
    \label{fig:dynamics}
\end{figure}

Figure~\ref{fig:dynamics}(a) shows the gradient of the classic and proposed \ac cGANs trained on Cityscapes~\cite{Cordts2016Cityscapes} label-to-image with and without \ac(see Section \ref{sec:lb-to-im}). The mean absolute value of the gradient is reported in order to demonstrate the stability of the training. Neither vanishing nor exploding gradient is observed for both models. 
Figure~\ref{fig:dynamics}(b) shows the training loss of the optimal discriminator trained as described in Section~\ref{sec:eval} for both models with the generator fixed at epoch $200$. Both models converge rapidly to $0$. Allowing the discriminator to converge for one epoch is enough to obtain the optimal discriminator with a fixed generator.

 \begin{figure}
    \centering
    \includegraphics[width=0.6\textwidth]{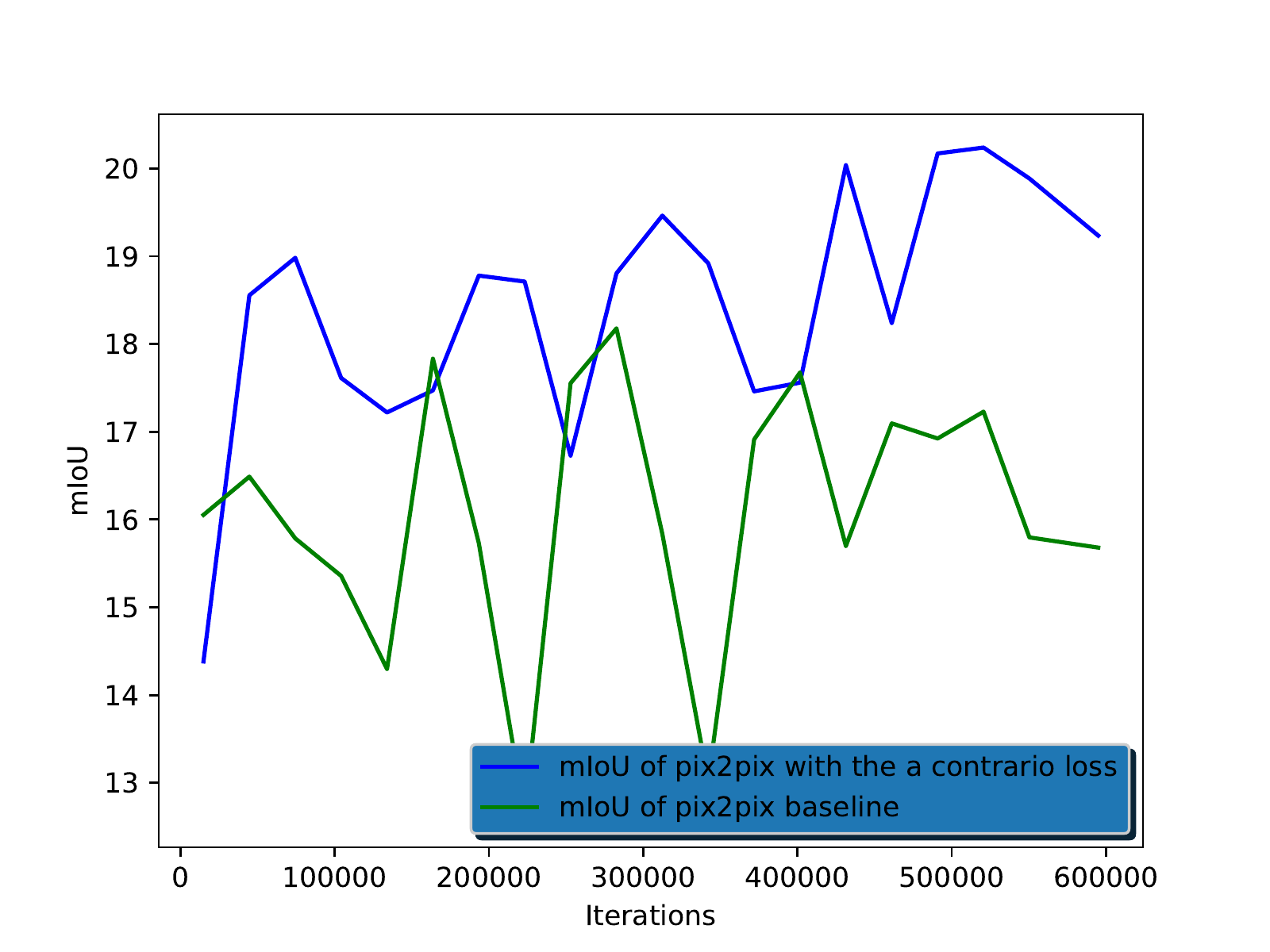}
    \caption{mIoU for the Cityscape image-to-label dataset throughout training. The proposed method consistently obtains more accurate results and finishes with a largely different score at the end of training $19.23$ versus for the baseline $15.97$.}
    \label{fig:city}
\end{figure}

\begin{figure*}
    \centering
    \includegraphics[width=0.8\textwidth]{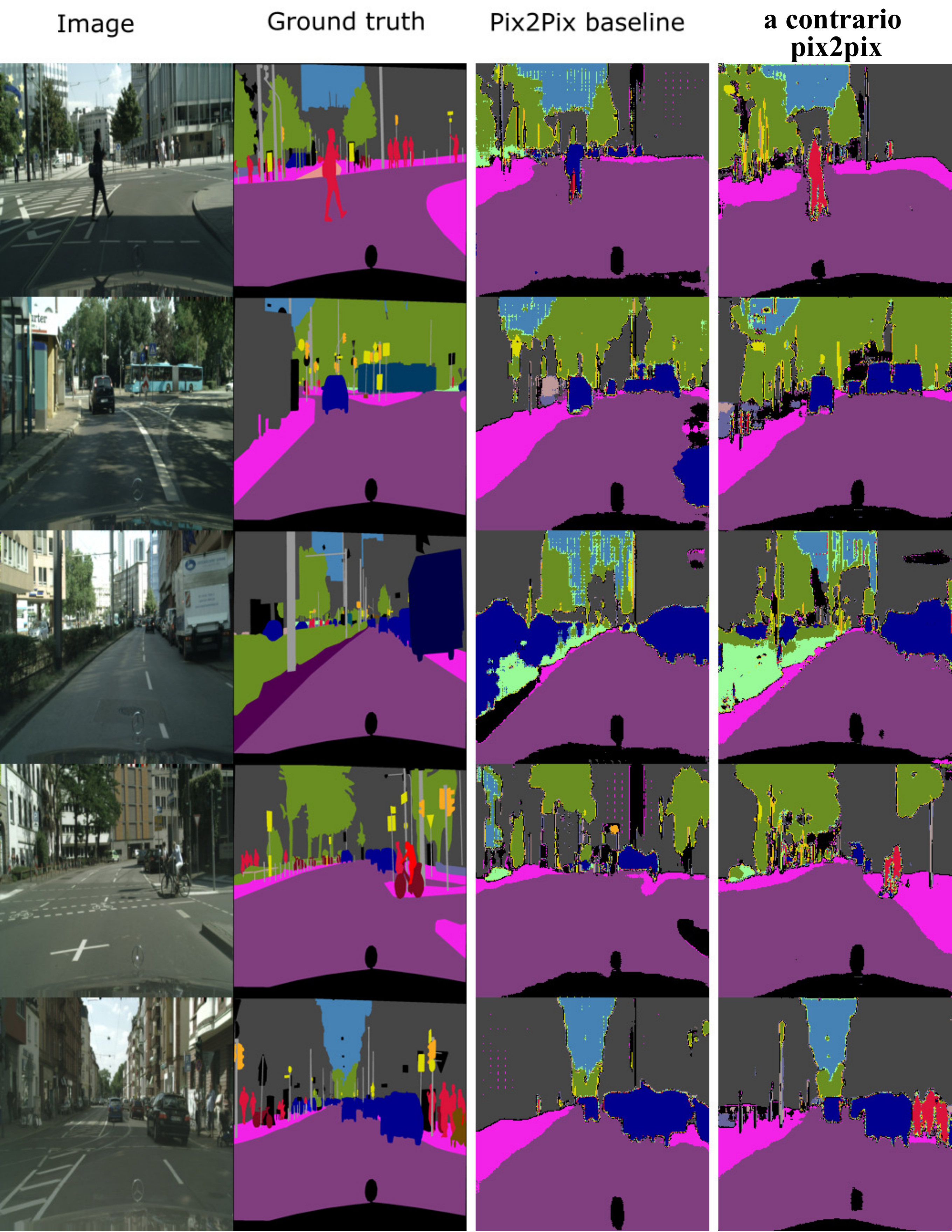}
    \caption{Qualitative results of Cityscape image-to-label task. It can be seen that the baseline model hallucinates objects. For instance, in the second row, the baseline hallucinates cars while the \ac cGAN segments the scene better. In the first row, the baseline wrongly classifies the pedestrian as a car. While training the model, the discriminator does not penalize the generator for these miss-classifications}
    \label{fig:seg_qualitative}
\end{figure*}

\begin{figure*}
    \centering
    \includegraphics[width=0.8\textwidth]{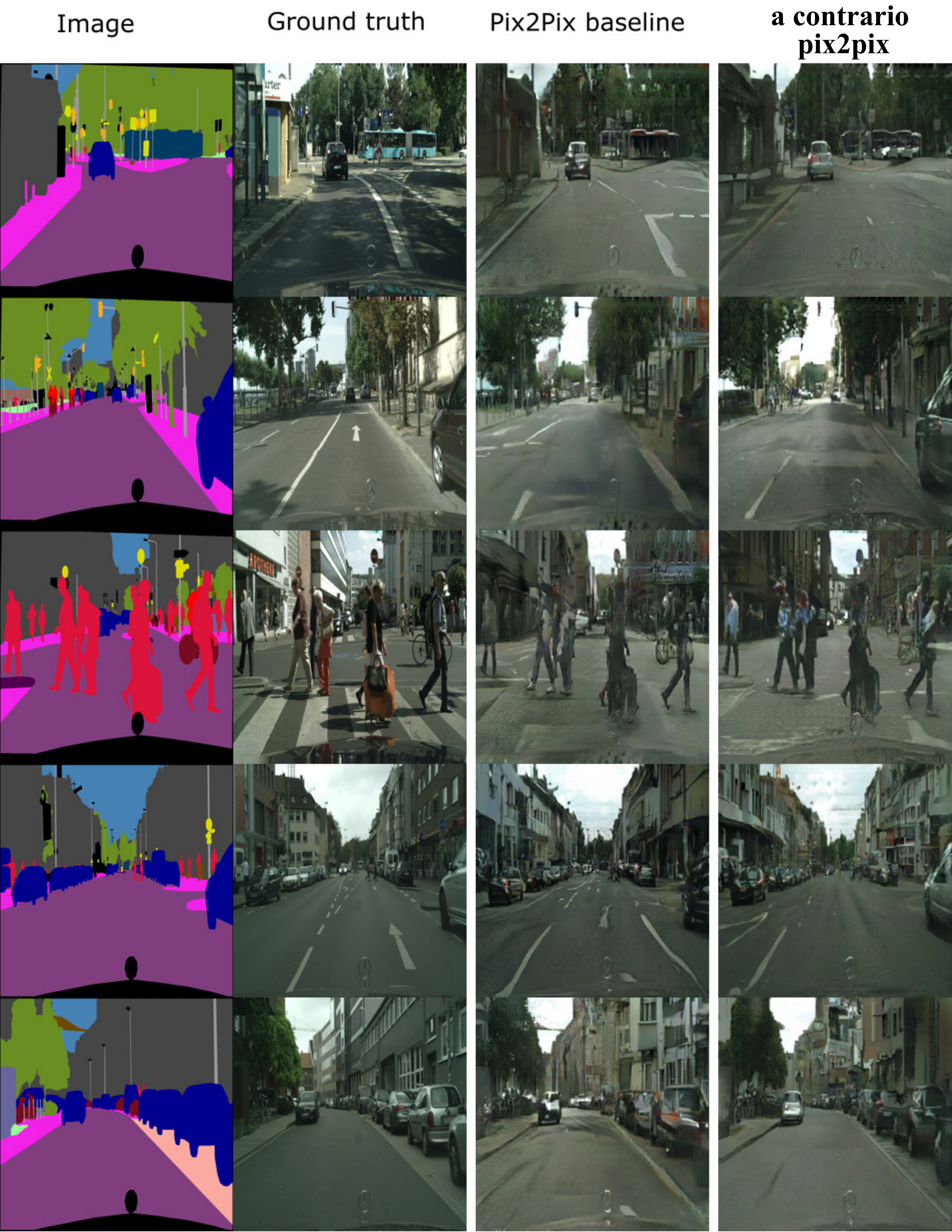}
    \caption{Qualitative results of Cityscapes label-to-image synthesis. In line with the quantitative results reported in Section \ref{sec:lb-to-im}, the qualitative results show better results for the \ac in comparison to the baseline. }
    \label{fig:cityscapes_qualitative}
\end{figure*}

\begin{figure}
    \centering
    \includegraphics[width=0.95\textwidth]{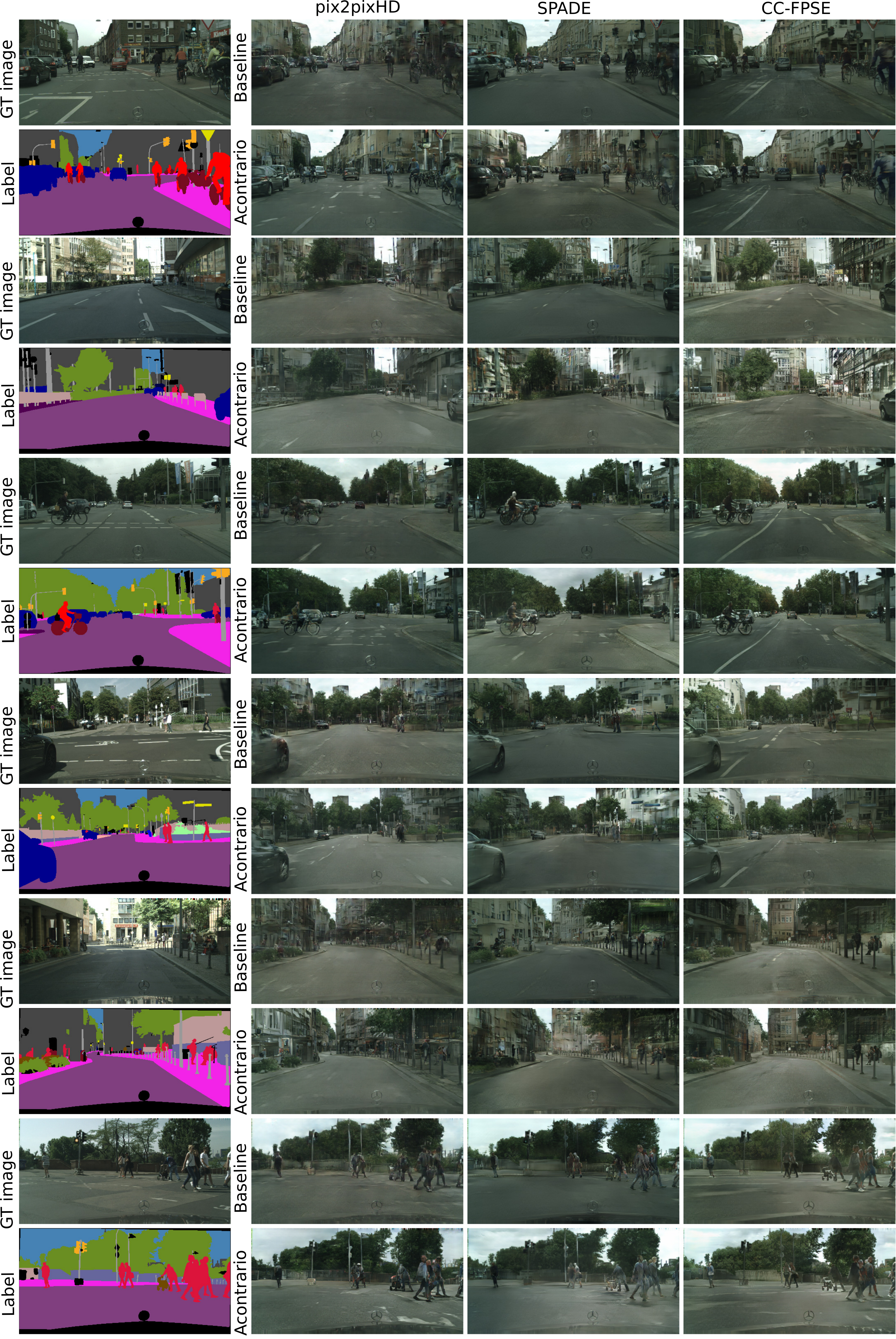}
    \caption{Qualitative comparison between different state-of-the-art methods for label-to-image trained and tested on Cityscapes\cite{Cordts2016Cityscapes} dataset. As observed, CC-FPSE baseline is the best baseline among classic cGAN. The \ac improves all the baseline and the best model among the 6 models is \ac CC-FPSE}
    \label{fig:state_of_art_comparison}
\end{figure}

\end{document}